\documentclass[11pt]{article}

\usepackage[preprint]{acl}

\usepackage{times}
\usepackage{latexsym}
\usepackage[T1]{fontenc}
\usepackage[utf8]{inputenc}
\usepackage{microtype}
\usepackage{inconsolata}
\usepackage{graphicx}
\usepackage{amsmath}
\usepackage{amssymb}
\usepackage{booktabs}
\usepackage{multirow}
\usepackage{placeins}

\usepackage{booktabs}
\usepackage{tabularx}
\usepackage{array}
\usepackage{xcolor}
\usepackage[skins,breakable]{tcolorbox}
\usepackage{soul}
\usepackage{enumitem}
\definecolor{findrule}{HTML}{5F8FA8}
\definecolor{findbg}{HTML}{EEF5F9}
\newcounter{finding}

\title{Decision Models as Judges for Hallucination Detection}

\author{
  \textbf{Haozheng Liu\textsuperscript{1}\thanks{Corresponding authors: Haozheng Liu and Lingyao Li.
\newline\hspace*{1.8em}
Emails: \texttt{hl6602@nyu.edu}, \texttt{lingyaoli@arizona.edu}.}},
  \textbf{Zhaoqian Xue\textsuperscript{2}},
  \textbf{Zhongyin Wang\textsuperscript{3}},
  \textbf{Junting Mu\textsuperscript{4}},
\\
  \textbf{Zihui Ma\textsuperscript{5,6}},
  \textbf{Shuya Feng\textsuperscript{7}},
  \textbf{Lingyao Li\textsuperscript{2}\footnotemark[2]}
\\[0.5ex]
  \normalsize
  \textsuperscript{1}New York University,
  \textsuperscript{2}University of Arizona,
\\
  \normalsize
  \textsuperscript{3}University of Illinois Urbana-Champaign,
  \textsuperscript{4}University of California, Irvine,
\\
  \normalsize
  \textsuperscript{5}Emory University,
  \textsuperscript{6}University of Maryland, College Park,
\\
  \normalsize
  \textsuperscript{7}University of Alabama at Birmingham
}

\begin{document}
\maketitle
\begin{abstract}
Large language models are widely used as judges of model outputs (LLM-as-a-Judge), but they are costly and can hallucinate or be overconfident. Decision models are a potential alternative. They return a predefined verdict with a probability for each label instead of free-form text and are much cheaper and faster. We test this promise on hallucination detection. We build binary benchmarks for five hallucination types (9,852 items from six datasets) and compare five decision models with the generative judge \textit{gpt-6-sol}. Performance depends strongly on the type: the best decision model matches \textit{gpt-6-sol} in balanced accuracy on context inconsistency and comes within 3 percentage points on factual errors, but trails it by 4--8 points on logical and instruction inconsistency. Decision-model probabilities are overconfident, though less so than the GPT-6 models, and separate correct from incorrect decisions only moderately. We also find that a confidence cascade of decision models improves on the best single model by 1.9--3.3 points on three types and matches or exceeds \textit{gpt-6-sol} at 1--3\% of its cost. Overall, decision models are a practical low-cost judge for some hallucination types but need to be validated before use.
\end{abstract}

\section{Introduction}
\label{sec:introduction}
\vspace{-2mm}

Large language models (LLMs) are widely used as judges of model outputs, an approach often called LLM-as-a-Judge~\cite{gu2026survey, zheng2023judging}. They rate open-ended responses~\cite{verga-etal-2024-juries,badshah-etal-2025-clev}, check whether responses follow instructions~\cite{wen-etal-2026-rewardbench,zeng2024evaluating}, and detect hallucinated content~\cite{li-etal-2023-halueval,tang-etal-2024-minicheck,xu-etal-2026-hallucination,manakul2023selfcheckgpt}. However, LLM judges are not fully reliable themselves. They can hallucinate, produce incorrect or unsupported judgments, and be overconfident in their judgments~\cite{kadavath-etal-2022-know,tian-etal-2023-just}.

Recently released decision models offer a different kind of judge. Models such as \textit{jev}~\cite{typesafe2026api}, \textit{d1}~\cite{liquidAI2026d1}, and \textit{Perplexity Decider}~\cite{nimble2026} do not generate free-form assessments but choose among predefined verdicts and return a probability for each. This design has two potential advantages. First, providers present these probabilities as a measure of certainty, so reliable probabilities can show when a judgment should not be trusted~\cite{typesafe2026api}. Second, decision models are much cheaper and faster than generative LLMs, which matters for large-scale evaluation~\cite{li-etal-2026-jev-judge,rao-callisonburch-2026-jev-rubric, ibrahim2026evaluating}.

Whether these advantages hold is still unclear. Recent studies of decision models as judges each examine a single model~\cite{li-etal-2026-jev-judge,rao-callisonburch-2026-jev-rubric}, leaving several questions open. First, judge performance varies across tasks~\cite{xu-etal-2025-context}, but how consistently decision models perform across error types is unknown. Second, whether their probabilities are calibrated and separate correct judgments from errors has merely been tested~\cite{xie2025uncertainty}. Third, multi-judge studies~\cite{verga-etal-2024-juries,ma-etal-2025-judging} have not tested ensembles composed solely of decision models for hallucination detection.

We address these gaps in the context of hallucination detection. First, judge reliability matters most here, since a judge that hallucinates while checking for hallucinations defeats its purpose. Second, hallucination types require different evidence (world knowledge, sources, the response itself, or the prompt), so they test whether decision models can generalize. We ask three research questions:

\begin{itemize}[leftmargin=*,nosep]
    \item \textbf{RQ1.} How accurately do decision models detect different hallucination types?
    \item \textbf{RQ2.} How reliable is their confidence?
    \item \textbf{RQ3.} Can ensembles of decision models improve detection accuracy?
\end{itemize}

We evaluate five decision models against the generative LLM \textit{gpt-6-sol} on 9,852 samples from six datasets based on five hallucination types \cite{huang-etal-2025-survey}. Our results show that decision models can match \textit{gpt-6-sol} on some hallucination types at a small fraction of its cost, and that most of them are only mildly overconfident. Confidence cascades of decision models further improve the accuracy of hallucination detection. Overall, our results show that decision models are a practical low-cost judge for some hallucination types but should be validated before use.

\section{Related Work}
\label{sec:literature}
\vspace{-2mm}

\begin{table*}[!t]
\centering
\footnotesize
\setlength{\tabcolsep}{5pt}
\renewcommand{\arraystretch}{1.2}
\begin{tabularx}{\textwidth}{@{} l >{\raggedright\arraybackslash}X >{\raggedright\arraybackslash}p{2.7cm} rrr @{}}
\toprule
\textbf{Hallucination type} & \textbf{Operational definition} & \textbf{Datasets} & \multicolumn{1}{c}{$H$} & \multicolumn{1}{c}{$N$} & \textbf{Total} \\
\midrule
Factual contradiction (FC)     & Claims contradict verifiable real-world facts through incorrect entities or relations. & FAVABench, HADTest & 344 & 1,277 & 1,621 \\
Factual fabrication (FF)       & Claims introduce fabricated, unverifiable, or unsupported facts or overclaims not established by real-world knowledge. & FAVABench, HADTest & 339 & 1,277 & 1,616 \\
Context inconsistency (CI)     & Claims conflict with, or are unsupported by, the supplied context or retrieved evidence. & RAGTruth & 943 & 1,732 & 2,675 \\
Logical inconsistency (LI)     & Claims or reasoning steps within the same response are mutually inconsistent. & ContraDoc, ROSCOE & 538 & 1,108 & 1,646 \\
Instruction inconsistency (II) & The response violates or fails to satisfy explicit instructions or constraints. & IF-RewardBench (English) & 1,600 & 694 & 2,294 \\
\midrule
\textbf{Total} & & & \textbf{3,764} & \textbf{6,088} & \textbf{9,852} \\
\bottomrule
\end{tabularx}
\caption{Hallucination-detection tasks and datasets. $H$ marks samples that contain the hallucination and $N$ do not.}

\label{tab:dataset_statistics}
\end{table*}

\noindent\textbf{LLM-as-a-Judge and Decision Models.}
LLM-as-a-Judge is widely used for automated evaluation. LLM judges can align with human evaluations \citep{min2023factscore} but also show biases such as position and verbosity bias \citep{zheng2023judging,liu2023g}, and because they generate text, they add cost and latency. In contrast, decision models provide a non-generative alternative by directly selecting from predefined labels and returning their associated probabilities. Recent work evaluates \textit{jev} on HaluEval factuality and final-answer adjudication \citep{li-etal-2026-jev-judge}, as well as on rubric criteria \citep{rao-callisonburch-2026-jev-rubric}, and reports lower cost and latency than generative LLM judges. \citet{ibrahim2026evaluating} compare several decision models on social-science annotation.

\par\vspace{2mm} \noindent\textbf{Hallucination detection.}
Detecting a hallucination requires evidence depending on its type \citep{huang-etal-2025-survey}. FAVABench and HADTest annotate factual errors \citep{mishra-etal-2024-fava,xu-etal-2026-hallucination}, RAGTruth annotates unsupported or contradictory content in retrieval-grounded responses \citep{niu-etal-2024-ragtruth}, ContraDoc and ROSCOE cover contradictions within documents and reasoning chains \citep{li-etal-2024-contradoc,golovneva-etal-2023-roscoe}, and IF-RewardBench covers instruction following \citep{wen-etal-2026-rewardbench}. AggreFact shows that factuality detector performance varies across datasets, summarizers, and error types \citep{tang-etal-2023-understanding}.

\par\vspace{2mm} \noindent\textbf{Confidence reliability.}
Returning probabilities does not make a judge's confidence reliable. Calibration asks whether probabilities match observed frequencies \citep{guo-etal-2017-calibration}, and error ranking asks whether confidence separates correct judgments from errors. Language-model confidence can reflect correctness, but this depends on the task and on how confidence is elicited \citep{kadavath-etal-2022-know,tian-etal-2023-just,xiong2024can}. For decision models, calibration varies by workload \citep{li-etal-2026-jev-judge,rao-callisonburch-2026-jev-rubric}, and high confidence can coexist with near-chance accuracy \citep{ibrahim2026evaluating}.

\par\vspace{2mm} \noindent\textbf{Judge ensembles and cascades.}
Combining judges helps when their errors differ. PoLL aggregates judgments from diverse generative models \citep{verga-etal-2024-juries}, and CLEV calls a third judge only when the first two disagree \citep{badshah-etal-2025-clev}, the idea behind our arbitration method. Confidence-based cascades pass uncertain items to another judge \citep{jung-etal-2025-trust}. JEV-to-generative cascades can match the generative judge at lower cost \citep{li-etal-2026-jev-judge}, although overlapping errors limit their gains on rubric tasks \citep{rao-callisonburch-2026-jev-rubric}.

Three areas remain underexplored. First, decision models have so far been evaluated one at a time or on tasks other than hallucination detection, so it is unclear how well they detect each hallucination type. Second, few studies have examined whether the confidence of decision models is reliable across models and hallucination types. Third, it has not been tested whether ensembles built only from decision models can improve hallucination detection while keeping their cost low. Our study aims to address these research gaps.

\section{Data and Methods}
\label{sec:methods}
\vspace{-2mm}

\subsection{Benchmark Composition}
\label{sec:benchmark-composition}


\noindent\textbf{Hallucination types.}
Hallucination takes several forms. Following the taxonomy of \citet{huang-etal-2025-survey}, we study five types (Table~\ref{tab:dataset_statistics}). Factual contradiction (FC) covers claims that contradict verifiable real-world facts. Factual fabrication (FF) covers unverifiable or unsupported overclaims. Context inconsistency (CI) covers claims that conflict with the supplied source material or are not supported by it. Logical inconsistency (LI) covers contradictions within a response or between its reasoning steps. Instruction inconsistency (II) covers violations of explicit instructions. We treat each type as a separate binary task, because one response can contain several types of error.

\begin{figure*}[!t]
  \centering
  \includegraphics[width=0.85\textwidth]{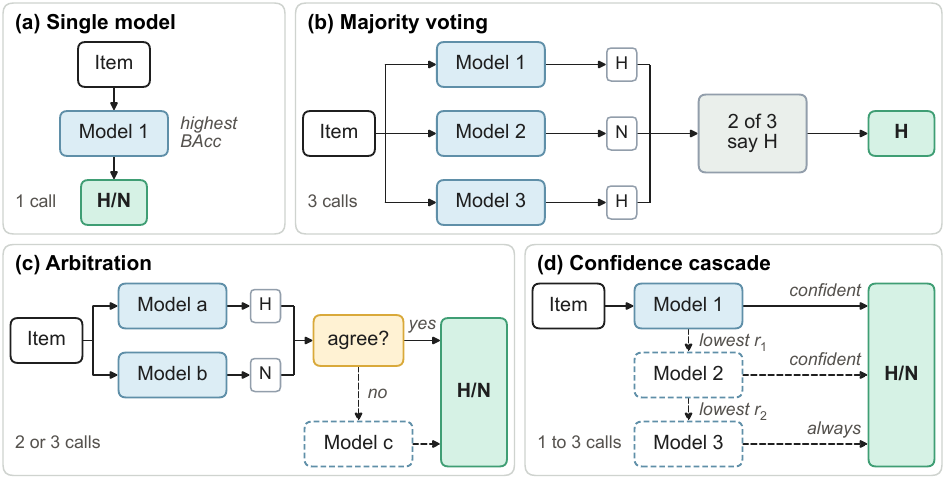}
  \caption{The single-model baseline and the three combination rules. Models~1--3: top-3 decision models by BAcc per type. Solid/dashed boxes: always/conditionally called. (c) The third model breaks ties between the two cheaper ones. (d) Each stage defers its least-confident $r_k$ share; the last stage decides the rest.}
  \vspace{-4mm}
  \label{fig:combination}
\end{figure*}

\par\vspace{2mm} \noindent\textbf{Benchmark datasets.}
We choose six benchmarks whose annotations mark the target error type explicitly, so each sample maps to a binary label without new annotation \citep{tang-etal-2023-understanding,tang-etal-2024-minicheck}. For FC and FF, we combine two datasets that annotate both types, given that FAVABench alone has fewer than 400 samples per type. FAVABench contains ChatGPT and Llama responses with human span-level error annotations \citep{mishra-etal-2024-fava}. HADTest contains responses into which GPT-4o injected a single factual error, validated by two annotators \citep{xu-etal-2026-hallucination}. For CI, RAGTruth contains responses from six LLMs to question answering, summarization, and data-to-text tasks, annotated for content that conflicts with the retrieved source or is unsupported by it \citep{niu-etal-2024-ragtruth}. For II, IF-RewardBench contains responses from 16 LLMs, each checked against a list of explicit constraints \citep{wen-etal-2026-rewardbench}. For LI, ContraDoc contains news, Wikipedia, and story documents with inserted self-contradictions \citep{li-etal-2024-contradoc}, and ROSCOE contains step-by-step reasoning annotated for contradictions \citep{golovneva-etal-2023-roscoe}.

An item is labeled $H$ if its source annotation marks the target error and $N$ if it marks the absence of that hallucination error. An $N$ sample may therefore still contain another type of hallucination. We keep each dataset's evaluation unit and do not rebalance the classes. Appendix~\ref{app:binary-mapping} details how we map each dataset's original annotations to the binary labels used in our evaluation.

\subsection{Evaluation Setup and Metrics}
\label{sec:protocol}

\noindent\textbf{Model selection.}
We select five decision models from five providers, including both proprietary and open-weight models. They are \textit{jev}, \textit{d1}, \textit{clef}, \textit{pplx}, and \textit{gpt-6-luna}. Details of these models are presented in Appendix~\ref{app:model-config}. For comparison, we also run the general-purpose generative LLM \textit{gpt-6-sol} on the same items, distinguishing it from \textit{gpt-6-luna}, which is included as a decision model because it directly outputs classification decisions. All six models are accessed through their hosted APIs. Latency is therefore end-to-end API response time and does not depend on local hardware.

\par\vspace{2mm} \noindent\textbf{Model input and output.}
For each hallucination type, all models receive the same input and the same $H/N$ criterion, based on the type's definition. The input consists of the response to be judged and, when available, the original instruction. For CI, it also includes the source context. The models never see the gold labels or the annotated error spans. Each model answers $H$ if the response contains the target error and $N$ otherwise, and returns probabilities $P(H)$ and $P(N)$ that sum to one. The answer is the label with the higher probability, and on an exact tie we keep the model's own answer. The decision models return these probabilities through their interfaces. For \textit{gpt-6-sol}, we ask for a structured verdict with its own probability estimates and no explanation.
Appendix~\ref{app:dm-prompts} provides the complete decision-model prompts.

\par\vspace{2mm} \noindent\textbf{Evaluation metrics.}
The classes are imbalanced (Table~\ref{tab:dataset_statistics}), so we compare each answer with the gold label and report balanced accuracy (BAcc), the mean recall over $H$ and $N$, and macro-F1, the mean F1 over the two classes. Differences between judges are tested with paired bootstrap over items (2,000 resamples) \citep{dror2018hitchhiker}. We also report the mean API latency per call and the estimated API cost. Cost is computed from recorded token usage at the providers' list prices at the time of evaluation.\footnote{Per million input tokens, \$0.042 for \textit{jev}, \$0.04 for \textit{d1}, \$0.24 for \textit{clef}, \$0.02 for \textit{pplx}, and \$0.10 for \textit{gpt-6-luna}. The decision models return a single choice token, so their output cost is negligible. For \textit{gpt-6-sol}, \$2 per million input tokens and \$10 per million output tokens.}

\subsection{Confidence Analysis}
\label{sec:conf-setup}

\noindent\textbf{Probabilities and confidence.}
For each item $i$, a decision model returns $p_i = P(\textsc{H} \mid x_i)$, the probability that the response contains the target hallucination type, and labels the response as hallucinated ($\hat{y}_i = 1$) when $p_i > 0.5$. Providers describe $p_i$ as an estimate of the probability that the predicate holds or as a calibrated score. The confidence of a decision is the probability of its predicted label,
\vspace{-2mm}
\begin{equation}
  c_i = \max(p_i,\, 1 - p_i) \in [0.5, 1].
\end{equation}
Under perfect calibration, predictions assigned confidence $q$ would be correct approximately $q$ of the time. We evaluate the extent to which this property holds in practice. 

\par\vspace{2mm} \noindent\textbf{Detection and calibration.}
Detection performance, i.e., how well $p_i$ ranks hallucinated above non-hallucinated responses, is measured by AUROC. At $\tau = 0.5$, we also report BAcc, MCC and the flag-rate bias $\Delta$ (defined as the share of responses flagged minus $\pi$ ($\Delta > 0$: over-flagging). Calibration is measured by the expected calibration error over $M = 10$ equal-width confidence bins \citep{guo-etal-2017-calibration,roelofs2022mitigating}:
\vspace{-2mm}
\begin{equation}
  \mathrm{ECE} = \sum_{m=1}^{M} \tfrac{|B_m|}{n}\,\bigl|\mathrm{acc}(B_m) - \mathrm{conf}(B_m)\bigr| .
\end{equation}
where $B_m$ is the set of predictions in confidence bin $m$. We also report the slope $b$ of the logistic model $\operatorname{logit}\Pr(y_i{=}1) = a + b\operatorname{logit}(p_i)$, where $b < 1$ indicates overconfidence \citep{calster2019calibration}.

\par\vspace{2mm} \noindent\textbf{Confidence as an error signal.}
A useful confidence score is expected to be higher for correct decisions ($\mathcal{C}$) than for errors ($\mathcal{E}$) \citep{hendrycks2016baseline,geifman2017selective}:
\vspace{-2mm}
\begin{equation}
  \text{conf-AUROC} = \Pr(c_i > c_j \mid i \in \mathcal{C},\, j \in \mathcal{E}),
\end{equation}
with ties counted as one half. A value of $0.5$ means that confidence cannot distinguish correct decisions from errors. We also report the share of errors made with high confidence ($c_i \geq 0.9$).

We report 95\% bootstrap intervals and compare models on the same items with paired bootstrap tests. A two-way ANOVA ($\eta^2$) separates cell and model effects, and Kendall's $W$ measures ranking consistency across cells.

\subsection{Model Ensembling Strategies}
\label{sec:combination}

Motivated by prior studies \cite{verga-etal-2024-juries, badshah-etal-2025-clev, jung-etal-2025-trust}, we evaluate three ensembling methods (Figure~\ref{fig:combination}) to assess whether they improve accuracy and at what additional cost. To isolate the effect of each method, we fix the model selection. For each type, all three methods combine the same three decision models, namely those with the highest BAcc on that type. 


\textbf{Majority voting.}
All three models judge the same item, and the label chosen by at least two of them is the verdict. Because every item gets all three calls, majority voting is the most expensive of the three methods. Voting gives every model the same weight, so two weaker models can outvote the strongest one. 

\textbf{Arbitration.}
When two models agree, the outcome is their shared label, and the third vote does not change it. When they disagree, the third vote decides the outcome. Arbitration therefore calls the two models first and calls the third only to resolve a disagreement \citep{badshah-etal-2025-clev}.

\textbf{Confidence cascade.}
A cascade lets the strongest model decide unless it is uncertain. Cascades usually order models from cheap to expensive to reduce calls to a costly final model \citep{chen2023frugalgpt,jung-etal-2025-trust}. We order the models by descending BAcc and ask whether consulting weaker models on the strongest model's least confident items improves BAcc~\citep{jitkrittum2023does}. At each of the first two stages, a model defers an item to the next model when its confidence $c_i$ falls below a threshold, and the third model decides all items that reach it. Because confidence distributions differ across models (Section~\ref{sec:rq2_results}), a common threshold would defer very different shares of items at different stages. We therefore set each threshold through a target deferral rate $r_k$, the share of items reaching stage $k$ that are deferred, so that each stage defers its least confident items without using gold labels. We evaluate $r_1, r_2 \in \{0, 0.1, \ldots, 0.5\}$: $r_1 = 0$ reduces the cascade to the strongest model, and above $0.5$ the strongest model would decide fewer than half of the items. As a control, we defer the same numbers of randomly chosen items.

\par\vspace{2mm} \noindent\textbf{Evaluation.}
For each hallucination type, we evaluate all $6 \times 6$  combinations of $r_1$ and $r_2$ and report the one with the highest BAcc. Because we choose this combination on the same items we evaluate, the result is an optimistic estimate of what a cascade can achieve\citep{cawley2010over}. To see how optimistic it is, we also choose the rates with five-fold cross-fitting, picking them on four folds and testing them on the fifth. The cost of an item is the total cost of all models called for it. For latency, we add up calls made one after another and take the slowest of calls made in parallel. We plot each method's BAcc against its cost, show the results of all 36 combinations separately, and use paired bootstrap tests to compare each method with its best single model and with \textit{gpt-6-sol}.

\section{Results}
\label{sec:results}
\vspace{-2mm}

\subsection{RQ1: How Do Decision Models Perform Across Hallucination Types?}
\label{sec:rq1_results}

\begin{table}[!t]
\centering
\footnotesize
\setlength{\tabcolsep}{2pt}
\renewcommand{\arraystretch}{1}
\begin{tabular}{@{}llrcrr@{}}
\toprule
Type (N) & Model & BAcc & Macro-F1 & Cost & Latency \\
\midrule

\multirow{6}{*}{\shortstack[c]{\textbf{FC}\\N=1{,}621}}
 & jev        & 75.01 & 74.96 & 0.0490 & 264.51 \\
 & d1         & 67.78 & 68.98 & 0.0295 & \textbf{224.15} \\
 & clef       & 72.12 & 73.95 & 0.2064 & 347.49 \\
 & pplx       & 78.93 & \textbf{78.48} & \textbf{0.0159} & 284.23 \\
 & gpt-6-luna & 77.75 & 68.36 & 0.0817 & 236.82 \\
 & gpt-6-sol  & \textbf{81.61} & 77.45 & 3.3320 & 1,825.76 \\
\midrule

\multirow{6}{*}{\shortstack[c]{\textbf{FF}\\N=1{,}616}}
 & jev        & 80.74 & 74.11 & 0.0570 & 239.63 \\
 & d1         & 70.51 & 74.40 & 0.0369 & \textbf{235.40} \\
 & clef       & 71.24 & 75.02 & 0.2498 & 354.56 \\
 & pplx       & 79.85 & \textbf{82.14} & \textbf{0.0195} & 290.10 \\
 & gpt-6-luna & \textbf{82.72} & 75.15 & 0.0999 & 236.64 \\
 & gpt-6-sol  & 82.05 & 81.36 & 3.5014 & 1,639.48 \\
\midrule

\multirow{6}{*}{\shortstack[c]{\textbf{CI}\\N=2{,}675}}
 & jev        & 82.53 & 82.09 & 0.2214 & 281.30 \\
 & d1         & 75.84 & 77.33 & 0.1797 & 319.34 \\
 & clef       & 68.21 & 69.38 & 1.0911 & 622.86 \\
 & pplx       & \textbf{82.66} & \textbf{84.55} & \textbf{0.0901} & 342.02 \\
 & gpt-6-luna & 64.04 & 53.01 & 0.4375 & \textbf{258.33} \\
 & gpt-6-sol  & 82.65 & 78.94 & 10.9699 & 2,002.29 \\
\midrule

\multirow{6}{*}{\shortstack[c]{\textbf{LI}\\N=1{,}646}}
 & jev        & 68.98 & 70.47 & 0.0733 & 284.50 \\
 & d1         & 58.46 & 56.95 & 0.0509 & 330.27 \\
 & clef       & 61.33 & 61.42 & 0.3398 & 515.09 \\
 & pplx       & 73.28 & 75.12 & \textbf{0.0270} & 284.83 \\
 & gpt-6-luna & 66.99 & 68.27 & 0.1356 & \textbf{281.61} \\
 & gpt-6-sol  & \textbf{77.14} & \textbf{78.32} & 4.4831 & 2,203.56 \\
\midrule

\multirow{6}{*}{\shortstack[c]{\textbf{II}\\N=2{,}294}}
 & jev        & 61.75 & 62.03 & 0.1403 & 276.94 \\
 & d1         & 62.50 & 62.02 & 0.1114 & 478.96 \\
 & clef       & 64.50 & 64.43 & 0.7021 & 593.57 \\
 & pplx       & 71.08 & 65.39 & \textbf{0.0566} & 306.38 \\
 & gpt-6-luna & 54.17 & 50.69 & 0.2846 & \textbf{263.23} \\
 & gpt-6-sol  & \textbf{78.88} & \textbf{78.43} & 8.5098 & 2,644.51 \\
\bottomrule
\end{tabular}
\caption{Single-judge results pooled by hallucination type. BAcc and Macro-F1 are percentages; Cost is the estimated total API cost for that type (USD), and Latency \ is mean per-call latency (ms).}
\vspace{-4mm}
\label{tab:model-performance}
\end{table}

\begin{figure}[!t]
  \centering
  \includegraphics[width=\columnwidth]{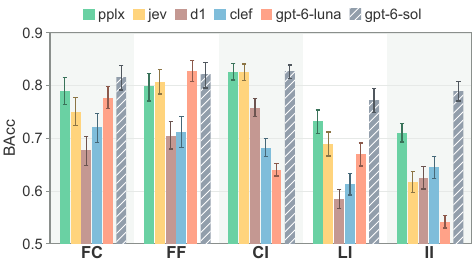}
  \caption{Judging performance by hallucination type. Bars show the pooled BAcc of each judge per type, with 95\% bootstrap confidence intervals. Hatched bars mark the generative model \textit{gpt-6-sol}.}
  \label{fig:rq1-bars}
  \vspace{-6mm}
\end{figure}

Table~\ref{tab:model-performance} and Figure~\ref{fig:rq1-bars} present the performance over each hallucination type. Appendix~\ref{app:rq1-details} reports their performance judge on every source dataset.

\textbf{Performance depends strongly on the type.}
\textit{gpt-6-sol} has the highest pooled BAcc on FC, LI, and II (Table~\ref{tab:model-performance}). On FF, \textit{gpt-6-luna} (82.7) edges it (82.0). On CI, \textit{pplx}, \textit{gpt-6-sol}, and \textit{jev} are within 0.2 points of one another at about 82.6. The gap is widest on II, where \textit{gpt-6-sol} reaches 78.9 against 71.1 for \textit{pplx} and 61.8 for \textit{jev}. Performance also varies within a model. \textit{jev} spans 20.8 points between II and CI. Type and source dataset are confounded, so we read these differences as type-specific without attributing them to the type alone.

\textbf{Similar scores hide different errors.}
The matched CI scores arise from different error patterns. \textit{pplx} rarely raises false alarms (false-positive rate 3.2\%) but misses many hallucinations (false-negative rate 31.5\%). \textit{gpt-6-sol} shows the reverse pattern, with a false-negative rate of 5.9\% and a false-positive rate of 28.8\%. \textit{jev} lies between them at 20.6\% and 14.4\%. The dominant error also changes with the task. \textit{jev} misses most positives on LI, with false-negative rates of 49\% on ContraDoc and 81\% on ROSCOE, but over-flags on II, where its false-positive rate is 56\%. BAcc alone therefore does not show whether a judge mostly misses hallucinations or mostly raises false alarms.

\textbf{Model advantages are concentrated in particular types.}
Relative to \textit{jev}, \textit{pplx} gains only 0.1 points on CI (95\% paired bootstrap interval $[-1.4, 1.6]$) and loses 0.9 points on FF ($[-3.4, 1.6]$), but it gains 3.9 points on FC ($[1.5, 6.4]$), 4.3 points on LI ($[2.2, 6.5]$), and 9.3 points on II ($[7.2, 11.5]$). The FC, LI, and II gains remain significant after Holm correction over the 25 comparisons of each judge with \textit{jev}. The ranking of decision models thus changes with the type.

\textbf{Advantages do not always hold across sources.}
On HADTest, all six judges score 12 to 18 points higher on FF than on FC. The same comparison on FAVABench shows no consistent gap (Appendix~\ref{app:rq1-details}). For \textit{pplx}, FF minus FC is $-1.4$ points on FAVABench (95\% bootstrap interval $[-8.5, 5.3]$) but $17.9$ points on HADTest ($[13.2, 22.9]$). An advantage on fabrication is therefore partly a property of the source. CI and II each have a single source, so their cross-source stability remains untested.

\textbf{Decision models are far cheaper and faster.}
Per item, the decision models cost between 0.5\% (\textit{pplx}) and 10\% (\textit{clef}) of \textit{gpt-6-sol}, and they answer in 0.22 to 0.62 seconds against 1.6 to 2.6 seconds (Table~\ref{tab:model-performance}). The cheapest model, \textit{pplx}, is also the most accurate decision model on four of the five types, and \textit{gpt-6-luna} is the most accurate on FF. The efficiency gain is thus available on every type. Whether it comes with the accuracy of \textit{gpt-6-sol} depends on the type. Additional analyses of error patterns and prediction overlap across models are provided in Appendix~\ref{app:error-patterns}.
\subsection{RQ2: How Reliable is Their Confidence?}
\label{sec:rq2_results}

\begin{figure}[!t]
  \centering
  \includegraphics[width=0.9\columnwidth]{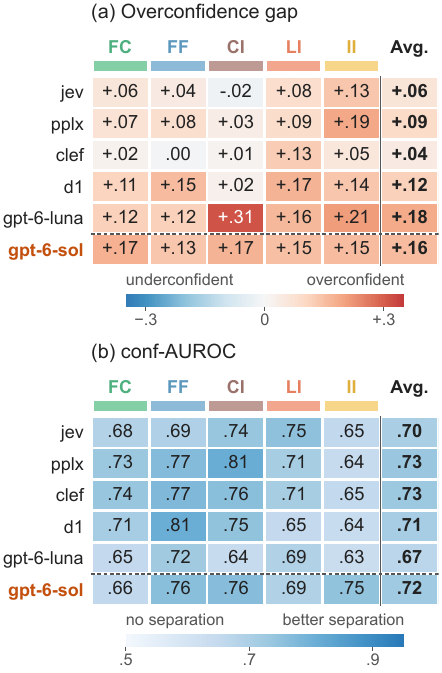}
  \caption{Confidence by model and hallucination type. (a) Overconfidence gap, i.e., mean confidence minus accuracy. (b) conf-AUROC; outlined values have a 95\% CI that includes $0.5$. }
  \vspace{-4mm}
  \label{fig:conf-type}
\end{figure}

\begin{figure*}[!t]
  \centering
  \includegraphics[width=0.95\textwidth]{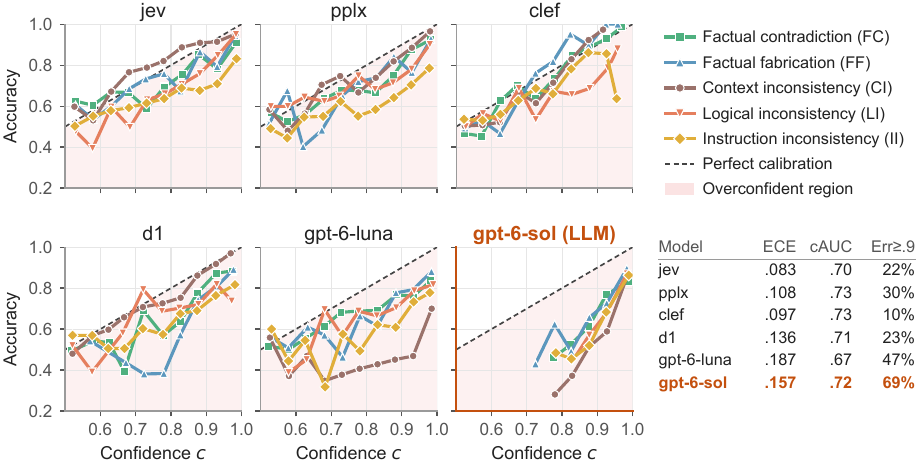}
  \vspace{-1mm}
  \caption{Reliability diagrams by hallucination type. Confidence is divided into 10 equal-width bins (bins with fewer than 25 items omitted); points below the diagonal indicate overconfidence. Right: ECE and conf-AUROC (cAUC) averaged over datasets, and the share of all errors (pooled over datasets) made with $c \geq 0.9$ (Err$\geq$.9).}
  \vspace{-4mm}
  \label{fig:conf-rel}
\end{figure*}

Figures \ref{fig:conf-type} and \ref{fig:conf-rel} present the confidence reliability of the six judges across hallucination types, including calibration and the ability to distinguish correct judgments from errors.

\textbf{Decision models are overconfident overall, but less so than the GPT-6 models.}
Every decision model has a positive average overconfidence gap, and the calibration slope is below 1 for 21 of the 25 model--type pairs (Figure~\ref{fig:conf-type}a). For \textit{jev}, \textit{pplx}, \textit{clef} and \textit{d1}, however, the overconfidence is mild. Mean confidence exceeds accuracy by $0.04$--$0.12$, and ECE ranges from $0.083$ to $0.136$. The two GPT-6 models are clearly more overconfident. \textit{gpt-6-luna} has a gap of $0.18$ and an ECE of $0.187$, and the comparison LLM \textit{gpt-6-sol} has a gap of $0.16$ and an ECE of $0.157$. Their median calibration slopes ($0.44$ and $0.36$) are also the lowest, compared with $0.54$--$0.58$ for \textit{jev}, \textit{d1} and \textit{pplx} and $0.96$ for \textit{clef}.

\textbf{Calibration depends more on the dataset than on the model.}
Across the decision models, the dataset accounts for 58\% of the variance in the overconfidence gap, and the model for 15\% (Figure~\ref{fig:conf-type}a). Averaged over the decision models, overconfidence is largest for II ($0.14$) and LI ($0.13$), and smallest for CI and factual contradiction FC ($0.07$ each). The reliability diagrams show the same pattern. At a given confidence level, accuracy differs by hallucination type (Figure~\ref{fig:conf-rel}). For example, the accuracy of \textit{gpt-6-luna} on CI stays below $0.50$ for all confidence levels between $0.55$ and $0.95$.

\textbf{Confidence is less reliable for instruction inconsistency (II) and logical inconsistency (LI).}
Averaged over the decision models, II has the largest overconfidence gap ($0.14$), the highest ECE ($0.145$) and the lowest conf-AUROC ($0.64$), followed by LI (gap $0.13$; ECE $0.136$; conf-AUROC $0.70$). FC is less overconfident (gap $0.07$; ECE $0.103$) but separates correct from incorrect decisions no better than LI (conf-AUROC $0.70$), whereas CI (gap $0.07$; ECE $0.093$; conf-AUROC $0.74$) and FF(gap $0.08$; ECE $0.134$; conf-AUROC $0.75$) are the most reliable. Figure~\ref{fig:conf-type} shows the same pattern. Among decisions made with $c \geq 0.9$, accuracy is lowest for II for \textit{jev} and \textit{pplx}, and for LI for \textit{clef} and \textit{d1} (Figure~\ref{fig:conf-rel}). Type-specific failures also occur for individual models. \textit{gpt-6-luna} is most overconfident on CI (gap $0.31$), where its accuracy stays below $0.50$.

\textbf{Confidence separates correct from incorrect decisions only moderately.} conf-AUROC measures how well confidence separates correct judgments from errors. A value of $0.5$ indicates performance no better than chance. Higher values indicate better separation between the two groups (Figure~\ref{fig:conf-type}b).
Averaged over the five hallucination types, conf-AUROC ranges from $0.67$ (\textit{gpt-6-luna}) to $0.73$ (\textit{pplx} and \textit{clef}). 
For each model and type, the 95\% confidence interval lies above $0.5$, so confidence always carries some information about correctness, but the separation remains moderate, especially for II.

\textbf{High confidence does not guarantee a correct decision.}
Errors can be made with high confidence ($c \geq 0.9$) (Figures~\ref{fig:conf-rel} and~\ref{fig:conf-dist}). Averaged over the five types, \textit{gpt-6-luna} assigns $c \geq 0.9$ to 65\% of its decisions, is correct on only 77\% of them, and makes 47\% of its errors at this level. For the other decision models, 10--30\% of errors are made at $c \geq 0.9$, with \textit{clef} lowest (10\%); \textit{clef} also assigns such high confidence to only 25\% of its decisions. The comparison model \textit{gpt-6-sol} is the most extreme case: it assigns $c \geq 0.9$ to 88\% of its decisions and makes 69\% of its errors at this level.


\begin{figure*}[!t]
  \centering
  \includegraphics[width=\textwidth]{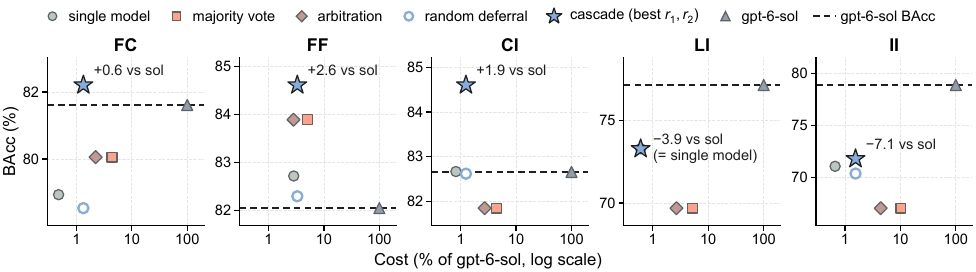}
  \caption{BAcc against estimated cost for each hallucination type. Cost is relative to \textit{gpt-6-sol} on a log scale, and the dashed line marks the BAcc of \textit{gpt-6-sol}. The cascade uses the deferral rates with the highest BAcc. The circle defers the same numbers of items at random.}
  \label{fig:quality-cost}
  \vspace{-4mm}
\end{figure*}

\subsection{RQ3: Can Ensembling Improve the Performance?}
\label{sec:rq3}

Figure~\ref{fig:quality-cost} presents the performance–cost trade-offs of the three ensembling methods. Deferral-rate sensitivity is reported in Appendix Figure~\ref{fig:cascade}.

\textbf{The cascade matches or exceeds GPT-6-sol on three types.}
At the best tested deferral rates, the confidence cascade achieves BAcc scores of 82.2, 84.6, and 84.6 on FC, FF, and CI, respectively---0.6, 2.6, and 1.9 points higher than \textit{gpt-6-sol}---at only 1.2--3.3\% of its estimated cost. The advantage over \textit{gpt-6-sol} is statistically significant on FF and CI, but not on FC. The cascade also improves on the strongest single member by 3.3, 1.9, and 2.0 points, respectively, with paired bootstrap 95\% intervals excluding zero. It makes 1.1--1.3 calls per item and has about one fifth of the latency of \textit{gpt-6-sol}. Selecting deferral rates by cross-fitting reduces BAcc by at most 0.52 points and preserves the gains over the strongest member.

\textbf{The cascade outperforms voting and arbitration.}
Majority voting gains about one BAcc point over the strongest single model on FC and FF, but falls below it on CI, LI, and II (Figure~\ref{fig:quality-cost}). Disagreement arbitration produces the same verdicts as voting while costing roughly half as much. Both rules are less accurate than the cascade on every type and, except on FF, also cost more. Unlike equal voting, the cascade consults weaker members only when the strongest member is uncertain.

\textbf{The best cascade configuration differs by type.}
\textit{pplx} leads four of the five cascades. On FF, \textit{gpt-6-luna} leads (Appendix Table~\ref{tab:cascade-settings}). The role of the later stages also varies. On CI, the best cascade never uses its third model ($r_2 = 0$), whereas on FF the second model defers about half of the samples it receives. On FC, FF, and CI, BAcc peaks at a first-stage deferral rate of 20--30\% and declines at higher rates (Appendix Figure~\ref{fig:cascade}). A single cascade configuration therefore does not transfer across types.

\textbf{The gap persists on LI and II.}
On LI, the best cascade defers no items and equals \textit{pplx} alone, 3.9 points below \textit{gpt-6-sol}. On II, the cascade improves BAcc by only 0.7 points over the strongest single model, \textit{pplx}, and remains 7.1 points below \textit{gpt-6-sol}. Other members do not improve on \textit{pplx} for its least confident items in balanced terms, consistent with the weaker confidence reliability on these two types (Section~\ref{sec:rq2_results}).

\section{Discussion and Conclusion}
\vspace{-2mm}
\label{sec:discussion}

We evaluate five decision models for hallucination detection. They match or approach the generative LLM judge on factual and context-related errors but lag on logical and instruction inconsistencies requiring multi-step reasoning or explicit constraint checking. We draw three takeaways.

\textbf{First, judges should be selected per hallucination type.} No decision model performs best across types, and similar BAcc can mask opposite error profiles: missed hallucinations or false alarms. The preferable judge thus depends on the target type and on which error is more costly, so false-negative and false-positive rates should be reported by type. Since type and source are partly confounded, judge selection should be validated on the target task.

\textbf{Second, confidence is useful but does not guarantee correctness.} Most decision models are less overconfident than \textit{gpt-6-luna} and \textit{gpt-6-sol}, but their confidence only moderately distinguishes correct from incorrect judgments, with frequent high-confidence errors on logical and instruction inconsistencies (Section~\ref{sec:rq2_results}). Confidence can guide requests for a second opinion but cannot establish correctness.

\textbf{Third, ensembling decision models helps when they are complementary on uncertain samples.} Cascades outperform their strongest member and match or exceed \textit{gpt-6-sol} on factual contradiction, factual fabrication, and context inconsistency. On logical and instruction inconsistency, the tested cascades offer limited gains, consistent with the observed error overlap (Appendix~\ref{app:error-patterns}). Since the best configuration varies by type, deferral rates should be validated on the target type to weigh accuracy gains against the cost of additional model calls.

At 0.5--10\% of the cost of \textit{gpt-6-sol}, decision models could support low-cost, type-aware cascades for factual and context errors, with LLM judges reserved for logical and instruction inconsistencies. Future work should disentangle type from data source, develop decision models that target reasoning- and instruction-level errors, and learn routing policies that decide when to escalate to an LLM judge based on detection accuracy and cost.

\section*{Limitations}
\label{sec:limitations}

This study presents several future directions. 

\textbf{Benchmark coverage and label consistency.}
Our benchmark covers five hallucination types across six datasets, but coverage is uneven: context and instruction inconsistency each rely on a single source. Differences in data construction, domain, and response length confound cross-dataset comparisons, leaving subgroup analyses observational. Mapping heterogeneous annotations to binary labels may also introduce mismatches because the source datasets define errors differently. Broader source coverage, annotation under a shared rubric, and comparisons matched on domain and response length would help address these limitations.

\textbf{Confidence comparability and prompt sensitivity.}
Confidence values are not fully comparable because decision models return label probabilities, whereas \textit{gpt-6-sol} reports its own estimates. We also use one prompt design per task, leaving prompt sensitivity untested. Comparing alternative prompts and, where supported, confidence-elicitation methods within each model would clarify their effects on accuracy and calibration.

\textbf{Generalization and deployment of cascades.}
We evaluate fixed model versions and assess cascades through offline replay and within-source cross-fitting, with member selection and ordering based on the full sample. Selecting all cascade components using training data alone, followed by evaluation on unseen domains and in live deployments, would better assess their accuracy, latency, and cost beyond the evaluated settings.

\section*{Statement of LLM Use}
\label{app:use_llm}

During the preparation of this paper, we used LLM-based tools (primarily Claude and ChatGPT) in two capacities. First, we used coding agents to assist with the development and refinement of the project codebase. Second, we used LLM chatbots to support language editing of manuscript drafts, including correcting grammatical errors, improving sentence structure, and enhancing the clarity of presentation. The authors take full responsibility for the content of this paper, including the code, analyses, and conclusions presented.

\newpage
\bibliography{reference}

@misc{nimble2026,
  author = {{Bespoke Labs} and Sathiamoorthy, Maheswaran},
  title = {Nimble},
  year = {2026},
  howpublished = {\url{https://github.com/bespokelabsai/nimble}},
  note = {Model: https://huggingface.co/bespokelabs/Bespoke-Nimble-9B}
}

@article{gu2026survey,
  title={A survey on {LLM}-as-a-judge},
  author={Gu, Jiawei and Jiang, Xuhui and Shi, Zhichao and Tan, Hexiang and Zhai, Xuehao and Xu, Chengjin and Li, Wei and Shen, Yinghan and Ma, Shengjie and Liu, Honghao and Wang, Saizhuo and Zhang, Kun and Lin, Zhouchi and Zhang, Bowen and Ni, Lionel and Gao, Wen and Wang, Yuanzhuo and Guo, Jian},
  journal={The Innovation},
  volume={7},
  number={6},
  year={2026},
  publisher={Elsevier}
}

@article{zheng2023judging,
  title={Judging {LLM}-as-a-Judge with {MT-Bench} and {Chatbot Arena}},
  author={Zheng, Lianmin and Chiang, Wei-Lin and Sheng, Ying and Zhuang, Siyuan and Wu, Zhanghao and Zhuang, Yonghao and Lin, Zi and Li, Zhuohan and Li, Dacheng and Xing, Eric P. and Zhang, Hao and Gonzalez, Joseph E. and Stoica, Ion},
  journal={Advances in Neural Information Processing Systems},
  volume={36},
  pages={46595--46623},
  year={2023}
}

@article{ibrahim2026evaluating,
  title={Evaluating decision models for text annotation in computational social science},
  author={Ibrahim, Hazem and Zaki, Yasir},
  journal={arXiv preprint arXiv:2609.24574},
  year={2026}
}

@article{liquidAI2026d1,
  author = {Liquid AI},
  title = {Introducing d1: The most capable decision model, now with vision},
  journal = {Liquid AI Blog},
  year = {2026},
  note = {www.liquid.ai/blog/d1-decision-model},
}

@misc{typesafe2026api,
  author = {{TypeSafe AI}},
  title = {{TypeSafe API}},
  year = {2026},
  note = {Version 0.2.0. Accessed September 24, 2026},
  url = {https://api.typesafe.ai/docs}
}

@inproceedings{xu-etal-2025-context,
  title = {Does Context Matter? {C}ontextual{J}udge{B}ench for Evaluating {LLM}-based Judges in Contextual Settings},
  author = {Xu, Austin and Bansal, Srijan and Ming, Yifei and Yavuz, Semih and Joty, Shafiq},
  booktitle = {Proceedings of the 63rd Annual Meeting of the Association for Computational Linguistics (Volume 1: Long Papers)},
  year = {2025},
  pages = {9541--9564},
  publisher = {Association for Computational Linguistics},
  doi = {10.18653/v1/2025.acl-long.470},
  url = {https://aclanthology.org/2025.acl-long.470/}
}

@inproceedings{xie2025uncertainty,
  title = {An Empirical Analysis of Uncertainty in Large Language Model Evaluations},
  author = {Xie, Qiujie and Li, Qingqiu and Yu, Zhuohao and Zhang, Yuejie and Zhang, Yue and Yang, Linyi},
  booktitle = {International Conference on Learning Representations},
  year = {2025},
  url = {https://proceedings.iclr.cc/paper_files/paper/2025/hash/82d3258eb58ceac31744a88005b7ddef-Abstract-Conference.html}
}

@inproceedings{ma-etal-2025-judging,
  title = {Judging with Many Minds: Do More Perspectives Mean Less Prejudice? On Bias Amplification and Resistance in Multi-Agent Based {LLM}-as-Judge},
  author = {Ma, Chiyu and Zhang, Enpei and Zhao, Yilun and Liu, Wenjun and Jia, Yaning and Qing, Peijun and Shi, Lin and Cohan, Arman and Yan, Yujun and Vosoughi, Soroush},
  booktitle = {Findings of the Association for Computational Linguistics: EMNLP 2025},
  year = {2025},
  pages = {17356--17392},
  publisher = {Association for Computational Linguistics},
  doi = {10.18653/v1/2025.findings-emnlp.941},
  url = {https://aclanthology.org/2025.findings-emnlp.941/}
}

@article{huang-etal-2025-survey,
  author = {Huang, Lei and Yu, Weijiang and Ma, Weitao and Zhong, Weihong and Feng, Zhangyin and Wang, Haotian and Chen, Qianglong and Peng, Weihua and Feng, Xiaocheng and Qin, Bing and Liu, Ting},
  title = {A Survey on Hallucination in Large Language Models: Principles, Taxonomy, Challenges, and Open Questions},
  journal = {ACM Transactions on Information Systems},
  year = {2025},
  volume = {43},
  number = {2},
  pages = {1--55},
  doi = {10.1145/3703155},
  url = {https://doi.org/10.1145/3703155},
}

@inproceedings{mishra-etal-2024-fava,
  author = {Mishra, Abhika and Asai, Akari and Balachandran, Vidhisha and Wang, Yizhong and Tsvetkov, Yulia and Neubig, Graham and Hajishirzi, Hannaneh},
  title = {Fine-grained Hallucination Detection and Editing for Language Models},
  booktitle = {First Conference on Language Modeling},
  year = {2024},
  url = {https://openreview.net/pdf?id=dJMTn3QOWO},
}

@inproceedings{xu-etal-2026-hallucination,
    title = "{HAD}: {HA}llucination Detection Language Models Based on a Comprehensive Hallucination Taxonomy",
    author = "Xu, Fan  and
      Hu, Xinyu  and
      Yu, Zhenghan  and
      Lin, Li  and
      Zhang, Xu  and
      Zhang, Yang  and
      Zhou, Wei  and
      Gu, Jinjie  and
      Wan, Xiaojun",
    editor = "Li, Yunyao  and
      Rehm, Georg  and
      Tu, Mei",
    booktitle = "Proceedings of the 64th Annual Meeting of the {A}ssociation for {C}omputational {L}inguistics (Volume 6: Industry Track)",
    month = jul,
    year = "2026",
    address = "San Diego, California, USA",
    publisher = "Association for Computational Linguistics",
    url = "https://aclanthology.org/2026.acl-industry.11/",
    doi = "10.18653/v1/2026.acl-industry.11",
    pages = "141--158",
    ISBN = "979-8-89176-394-4"
}

@inproceedings{niu-etal-2024-ragtruth,
    title = "{RAGT}ruth: A Hallucination Corpus for Developing Trustworthy Retrieval-Augmented Language Models",
    author = "Niu, Cheng  and
      Wu, Yuanhao  and
      Zhu, Juno  and
      Xu, Siliang  and
      Shum, KaShun  and
      Zhong, Randy  and
      Song, Juntong  and
      Zhang, Tong",
    editor = "Ku, Lun-Wei  and
      Martins, Andre  and
      Srikumar, Vivek",
    booktitle = "Proceedings of the 62nd Annual Meeting of the Association for Computational Linguistics (Volume 1: Long Papers)",
    month = aug,
    year = "2024",
    address = "Bangkok, Thailand",
    publisher = "Association for Computational Linguistics",
    url = "https://aclanthology.org/2024.acl-long.585/",
    doi = "10.18653/v1/2024.acl-long.585",
    pages = "10862--10878"
}

@inproceedings{tang-etal-2023-understanding,
    title = "Understanding Factual Errors in Summarization: Errors, Summarizers, Datasets, Error Detectors",
    author = "Tang, Liyan  and
      Goyal, Tanya  and
      Fabbri, Alex  and
      Laban, Philippe  and
      Xu, Jiacheng  and
      Yavuz, Semih  and
      Kryscinski, Wojciech  and
      Rousseau, Justin  and
      Durrett, Greg",
    editor = "Rogers, Anna  and
      Boyd-Graber, Jordan  and
      Okazaki, Naoaki",
    booktitle = "Proceedings of the 61st Annual Meeting of the Association for Computational Linguistics (Volume 1: Long Papers)",
    month = jul,
    year = "2023",
    address = "Toronto, Canada",
    publisher = "Association for Computational Linguistics",
    url = "https://aclanthology.org/2023.acl-long.650/",
    doi = "10.18653/v1/2023.acl-long.650",
    pages = "11626--11644"
}

@inproceedings{tang-etal-2024-minicheck,
    title = "{M}ini{C}heck: Efficient Fact-Checking of {LLM}s on Grounding Documents",
    author = "Tang, Liyan  and
      Laban, Philippe  and
      Durrett, Greg",
    editor = "Al-Onaizan, Yaser  and
      Bansal, Mohit  and
      Chen, Yun-Nung",
    booktitle = "Proceedings of the 2024 Conference on Empirical Methods in Natural Language Processing",
    month = nov,
    year = "2024",
    address = "Miami, Florida, USA",
    publisher = "Association for Computational Linguistics",
    url = "https://aclanthology.org/2024.emnlp-main.499/",
    doi = "10.18653/v1/2024.emnlp-main.499",
    pages = "8818--8847"
}

@misc{li-etal-2026-jev-judge,
  author = {Li, Yubo and Miao, Yidi and Krishnan, Ramayya and Padman, Rema},
  title = {{JEV}-as-a-Judge: Accept When Confident, Escalate When Unsure},
  year = {2026},
  howpublished = {arXiv preprint arXiv:2609.26550},
  eprint = {2609.26550},
  archivePrefix = {arXiv},
  primaryClass = {cs.AI},
  note = {Version 4, 6 October 2026},
  url = {https://arxiv.org/abs/2609.26550v4},
  doi = {10.48550/arXiv.2609.26550},
}

@misc{rao-callisonburch-2026-jev-rubric,
  author = {Rao, Delip and Callison-Burch, Chris},
  title = {{JEV} vs. {LLM}s as Rubric Judges: Cheaper, Faster, and Wrong in the Same Places},
  year = {2026},
  howpublished = {arXiv preprint arXiv:2609.29769},
  eprint = {2609.29769},
  archivePrefix = {arXiv},
  primaryClass = {cs.CL},
  note = {Version 2, 28 September 2026},
  url = {https://arxiv.org/abs/2609.29769v2},
  doi = {10.48550/arXiv.2609.29769},
}

@inproceedings{guo-etal-2017-calibration,
  author = {Guo, Chuan and Pleiss, Geoff and Sun, Yu and Weinberger, Kilian Q.},
  title = {On Calibration of Modern Neural Networks},
  booktitle = {Proceedings of the 34th International Conference on Machine Learning},
  year = {2017},
  volume = {70},
  series = {Proceedings of Machine Learning Research},
  pages = {1321--1330},
  publisher = {PMLR},
  url = {https://proceedings.mlr.press/v70/guo17a.html},
}

@misc{kadavath-etal-2022-know,
  author = {Kadavath, Saurav and Conerly, Tom and Askell, Amanda and Henighan, Tom and Drain, Dawn and Perez, Ethan and Schiefer, Nicholas and Hatfield-Dodds, Zac and DasSarma, Nova and Tran-Johnson, Eli and Johnston, Scott and El-Showk, Sheer and Jones, Andy and Elhage, Nelson and Hume, Tristan and Chen, Anna and Bai, Yuntao and Bowman, Sam and Fort, Stanislav and Ganguli, Deep and Hernandez, Danny and Jacobson, Josh and Kernion, Jackson and Kravec, Shauna and Lovitt, Liane and Ndousse, Kamal and Olsson, Catherine and Ringer, Sam and Amodei, Dario and Brown, Tom and Clark, Jack and Joseph, Nicholas and Mann, Ben and McCandlish, Sam and Olah, Chris and Kaplan, Jared},
  title = {Language Models (Mostly) Know What They Know},
  year = {2022},
  howpublished = {arXiv preprint arXiv:2207.05221},
  eprint = {2207.05221},
  archivePrefix = {arXiv},
  primaryClass = {cs.CL},
  url = {https://arxiv.org/abs/2207.05221},
  doi = {10.48550/arXiv.2207.05221},
}

@misc{verga-etal-2024-juries,
  author = {Verga, Pat and Hofst{\"a}tter, Sebastian and Althammer, Sophia and Su, Yixuan and Piktus, Aleksandra and Arkhangorodsky, Arkady and Xu, Minjie and White, Naomi and Lewis, Patrick},
  title = {Replacing Judges with Juries: Evaluating {LLM} Generations with a Panel of Diverse Models},
  year = {2024},
  howpublished = {arXiv preprint arXiv:2404.18796},
  eprint = {2404.18796},
  archivePrefix = {arXiv},
  primaryClass = {cs.CL},
  url = {https://arxiv.org/abs/2404.18796},
  doi = {10.48550/arXiv.2404.18796},
}

@inproceedings{tian-etal-2023-just,
    title = "Just Ask for Calibration: Strategies for Eliciting Calibrated Confidence Scores from Language Models Fine-Tuned with Human Feedback",
    author = "Tian, Katherine  and
      Mitchell, Eric  and
      Zhou, Allan  and
      Sharma, Archit  and
      Rafailov, Rafael  and
      Yao, Huaxiu  and
      Finn, Chelsea  and
      Manning, Christopher",
    editor = "Bouamor, Houda  and
      Pino, Juan  and
      Bali, Kalika",
    booktitle = "Proceedings of the 2023 Conference on Empirical Methods in Natural Language Processing",
    month = dec,
    year = "2023",
    address = "Singapore",
    publisher = "Association for Computational Linguistics",
    url = "https://aclanthology.org/2023.emnlp-main.330/",
    doi = "10.18653/v1/2023.emnlp-main.330",
    pages = "5433--5442"
}

@inproceedings{badshah-etal-2025-clev,
    title = "{CLEV}: {LLM}-Based Evaluation Through Lightweight Efficient Voting for Free-Form Question-Answering",
    author = "Badshah, Sher  and
      Moustafa, Moamen  and
      Sajjad, Hassan",
    editor = "Inui, Kentaro  and
      Sakti, Sakriani  and
      Wang, Haofen  and
      Wong, Derek F.  and
      Bhattacharyya, Pushpak  and
      Banerjee, Biplab  and
      Ekbal, Asif  and
      Chakraborty, Tanmoy  and
      Singh, Dhirendra Pratap",
    booktitle = "Proceedings of the 14th International Joint Conference on Natural Language Processing and the 4th Conference of the Asia-Pacific Chapter of the Association for Computational Linguistics",
    month = dec,
    year = "2025",
    address = "Mumbai, India",
    publisher = "The Asian Federation of Natural Language Processing and The Association for Computational Linguistics",
    url = "https://aclanthology.org/2025.findings-ijcnlp.93/",
    doi = "10.18653/v1/2025.findings-ijcnlp.93",
    pages = "1513--1531",
    ISBN = "979-8-89176-303-6"
}

@inproceedings{jung-etal-2025-trust,
 author = {Jung, Jaehun and Brahman, Faeze and Choi, Yejin},
 booktitle = {International Conference on Learning Representations},
 editor = {Y. Yue and A. Garg and N. Peng and F. Sha and R. Yu},
 pages = {3101--3125},
 title = {Trust or Escalate: {LLM} Judges with Provable Guarantees for Human Agreement},
 url = {https://proceedings.iclr.cc/paper_files/paper/2025/file/08dabd5345b37fffcbe335bd578b15a0-Paper-Conference.pdf},
 volume = {2025},
 year = {2025}
}

@inproceedings{li-etal-2024-contradoc,
    title = "{C}ontra{D}oc: Understanding Self-Contradictions in Documents with Large Language Models",
    author = "Li, Jierui  and
      Raheja, Vipul  and
      Kumar, Dhruv",
    editor = "Duh, Kevin  and
      Gomez, Helena  and
      Bethard, Steven",
    booktitle = "Proceedings of the 2024 Conference of the North American Chapter of the Association for Computational Linguistics: Human Language Technologies (Volume 1: Long Papers)",
    month = jun,
    year = "2024",
    address = "Mexico City, Mexico",
    publisher = "Association for Computational Linguistics",
    url = "https://aclanthology.org/2024.naacl-long.362/",
    doi = "10.18653/v1/2024.naacl-long.362",
    pages = "6509--6523"
}

@inproceedings{wen-etal-2026-rewardbench,
    title = "{IF}-{R}eward{B}ench: Benchmarking Judge Models for Instruction-Following Evaluation",
    author = "Wen, Bosi  and
      Niu, Yilin  and
      Wang, Cunxiang  and
      Ling, Xiaoying  and
      Zhang, Ying  and
      Ke, Pei  and
      Wang, Hongning  and
      Huang, Minlie",
    editor = "Liakata, Maria  and
      Moreira, Viviane P.  and
      Zhang, Jiajun  and
      Jurgens, David",
    booktitle = "Proceedings of the 64th Annual Meeting of the {A}ssociation for {C}omputational {L}inguistics (Volume 1: Long Papers)",
    month = jul,
    year = "2026",
    address = "San Diego, California, United States",
    publisher = "Association for Computational Linguistics",
    url = "https://aclanthology.org/2026.acl-long.1092/",
    doi = "10.18653/v1/2026.acl-long.1092",
    pages = "23816--23843",
    ISBN = "979-8-89176-390-6"
}

@inproceedings{li-etal-2023-halueval,
    title = "{H}alu{E}val: A Large-Scale Hallucination Evaluation Benchmark for Large Language Models",
    author = "Li, Junyi  and
      Cheng, Xiaoxue  and
      Zhao, Wayne Xin  and
      Nie, Jian-Yun  and
      Wen, Ji-Rong",
    editor = "Bouamor, Houda  and
      Pino, Juan  and
      Bali, Kalika",
    booktitle = "Proceedings of the 2023 Conference on Empirical Methods in Natural Language Processing",
    month = dec,
    year = "2023",
    address = "Singapore",
    publisher = "Association for Computational Linguistics",
    url = "https://aclanthology.org/2023.emnlp-main.397/",
    doi = "10.18653/v1/2023.emnlp-main.397",
    pages = "6449--6464"
}

@inproceedings{golovneva-etal-2023-roscoe,
  author = {Golovneva, Olga and Chen, Moya and Poff, Spencer and Corredor, Martin and Zettlemoyer, Luke and Fazel-Zarandi, Maryam and Celikyilmaz, Asli},
  title = {{ROSCOE}: A Suite of Metrics for Scoring Step-by-Step Reasoning},
  booktitle = {The Eleventh International Conference on Learning Representations},
  year = {2023},
  url = {https://arxiv.org/abs/2212.07919v2},
  note = {ICLR 2023; author version arXiv:2212.07919v2},
}

@article{chen2023frugalgpt,
  title   = {{FrugalGPT}: How to Use Large Language Models While Reducing Cost and Improving Performance},
  author  = {Chen, Lingjiao and Zaharia, Matei and Zou, James},
  journal = {Transactions on Machine Learning Research},
  year    = {2024},
  note    = {arXiv:2305.05176},
  url     = {https://arxiv.org/abs/2305.05176},
}

@article{cawley2010over,
  title={On over-fitting in model selection and subsequent selection bias in performance evaluation},
  author={Cawley, Gavin C and Talbot, Nicola LC},
  journal={The Journal of Machine Learning Research},
  volume={11},
  pages={2079--2107},
  year={2010},
  publisher={JMLR. org}
}

@inproceedings{liu2023g,
  title={G-eval: NLG evaluation using gpt-4 with better human alignment},
  author={Liu, Yang and Iter, Dan and Xu, Yichong and Wang, Shuohang and Xu, Ruochen and Zhu, Chenguang},
  booktitle={Proceedings of the 2023 conference on empirical methods in natural language processing},
  pages={2511--2522},
  year={2023}
}

@article{jitkrittum2023does,
  title={When does confidence-based cascade deferral suffice?, 2024},
  author={Jitkrittum, Wittawat and Gupta, Neha and Menon, Aditya Krishna and Narasimhan, Harikrishna and Rawat, Ankit Singh and Kumar, Sanjiv},
  journal={URL https://arxiv. org/abs/2307.02764},
  year={2023}
}

@article{hendrycks2016baseline,
  title={A baseline for detecting misclassified and out-of-distribution examples in neural networks},
  author={Hendrycks, Dan and Gimpel, Kevin},
  journal={arXiv preprint arXiv:1610.02136},
  year={2016}
}

@article{calster2019calibration,
  title={Calibration: the Achilles heel of predictive analytics},
  author={Calster, B},
  journal={BMC Med},
  volume={17},
  number={1},
  pages={1},
  year={2019}
}

@inproceedings{roelofs2022mitigating,
  title={Mitigating bias in calibration error estimation},
  author={Roelofs, Rebecca and Cain, Nicholas and Shlens, Jonathon and Mozer, Michael C},
  booktitle={International Conference on Artificial Intelligence and Statistics},
  pages={4036--4054},
  year={2022},
  organization={PMLR}
}

@inproceedings{dror2018hitchhiker,
  title={The hitchhiker’s guide to testing statistical significance in natural language processing},
  author={Dror, Rotem and Baumer, Gili and Shlomov, Segev and Reichart, Roi},
  booktitle={Proceedings of the 56th annual meeting of the association for computational linguistics (volume 1: Long papers)},
  pages={1383--1392},
  year={2018}
}

@article{xiong2024can,
  title={Can LLMs express their uncertainty},
  author={Xiong, Miao and Hu, Zhiyuan and Lu, Xinyang and Li, Yifei and Fu, Jie and He, Junxian and Hooi, Bryan},
  journal={An empirical evaluation of confidence elicitation in LLMs. arXiv},
  volume={2306},
  year={2024}
}

@article{geifman2017selective,
  title={Selective classification for deep neural networks},
  author={Geifman, Yonatan and El-Yaniv, Ran},
  journal={Advances in neural information processing systems},
  volume={30},
  year={2017}
}

@inproceedings{manakul2023selfcheckgpt,
  title={Selfcheckgpt: Zero-resource black-box hallucination detection for generative large language models},
  author={Manakul, Potsawee and Liusie, Adian and Gales, Mark},
  booktitle={Proceedings of the 2023 conference on empirical methods in natural language processing},
  pages={9004--9017},
  year={2023}
}

@inproceedings{min2023factscore,
  title={FActScore: Fine-grained atomic evaluation of factual precision in long form text generation},
  author={Min, Sewon and Krishna, Kalpesh and Lyu, Xinxi and Lewis, Mike and Yih, Wen-tau and Koh, Pang and Iyyer, Mohit and Zettlemoyer, Luke and Hajishirzi, Hannaneh},
  booktitle={Proceedings of the 2023 conference on empirical methods in natural language processing},
  pages={12076--12100},
  year={2023}
}

@inproceedings{zeng2024evaluating,
  title={Evaluating large language models at evaluating instruction following},
  author={Zeng, Zhiyuan and Yu, Jiatong and Gao, Tianyu and Meng, Yu and Goyal, Tanya and Chen, Danqi},
  booktitle={International Conference on Learning Representations},
  volume={2024},
  pages={40193--40219},
  year={2024}
}
\clearpage
\appendix
\raggedbottom
\widowpenalty=10000
\clubpenalty=10000
\captionsetup{skip=5pt,hypcap=false,justification=justified,singlelinecheck=false}
\captionsetup[table]{position=bottom,aboveskip=5pt,belowskip=0pt}
\captionsetup[figure]{position=bottom,aboveskip=5pt,belowskip=0pt}

\makeatletter
\setlength{\@fptop}{0pt}
\setlength{\@dblfptop}{0pt}
\setlength{\@fpsep}{14pt plus 2pt minus 2pt}
\setlength{\@dblfpsep}{14pt plus 2pt minus 2pt}
\makeatother
\setlength{\floatsep}{10pt plus 2pt minus 2pt}
\setlength{\dblfloatsep}{14pt plus 2pt minus 2pt}
\setlength{\textfloatsep}{12pt plus 2pt minus 2pt}
\setlength{\dbltextfloatsep}{12pt plus 2pt minus 2pt}
\renewcommand{\dbltopfraction}{0.85}
\renewcommand{\dblfloatpagefraction}{0.8}
\setcounter{dbltopnumber}{3}
\makeatletter
\@secpenalty=0
\makeatother

\newcommand{\apptablestyle}{%
  \footnotesize
  \setlength{\tabcolsep}{4pt}%
  \renewcommand{\arraystretch}{1.0}%
  \setlength{\aboverulesep}{0.35ex}%
  \setlength{\belowrulesep}{0.45ex}}

\definecolor{promptbar}{HTML}{3A4E63}
\newtcolorbox{promptbox}[2][]{enhanced, breakable, colback=white, colframe=promptbar,
  colbacktitle=promptbar, coltitle=white, fonttitle=\small\bfseries, fontupper=\footnotesize\ttfamily,
  title={#2}, boxrule=0.6pt, arc=1.2pt, left=4pt, right=4pt, top=3pt, bottom=3pt,
  toptitle=2pt, bottomtitle=2pt, before skip=8pt, after skip=8pt,
  before upper={\raggedright\setlength{\parskip}{3pt}}, #1}
\makeatletter
\def\dm@statelabel{State}
\newcommand{\dmfield}[2]{\def\dm@l{#1}%
  \ifx\dm@l\dm@statelabel
    \par\noindent\textbf{#1:}\par\nobreak
    {\def\allowbreak{\par\noindent}\noindent #2\par}%
  \else
    \par\noindent\textbf{#1:}\enspace #2\par
  \fi}
\makeatother

\makeatletter
\newcommand{\appsection}[1]{%
  \section{#1}%
  \addcontentsline{apx}{section}{\protect\numberline{\thesection}#1}}
\newcommand{\appsubsection}[1]{%
  \subsection{#1}%
  \addcontentsline{apx}{subsection}{\protect\numberline{\thesubsection}#1}}
\newcommand{\appendixcontents}{%
  \section*{Appendix Contents}%
  \begingroup
  \renewcommand{\l@section}{\@dottedtocline{1}{0em}{1.6em}}%
  \renewcommand{\l@subsection}{\@dottedtocline{2}{1.6em}{2.8em}}%
  \@starttoc{apx}%
  \endgroup\par\vspace{6pt}}
\makeatother

\twocolumn[{%
\begin{minipage}{\textwidth}
\appendixcontents
\appsection{Benchmark Construction, Labels, and Prompts}
\label{app:label-mapping}
\appsubsection{Hallucination Taxonomy and Examples}
\label{app:taxonomy}
\par\vspace{4pt}
\begingroup
\centering
\apptablestyle
\begin{tabularx}{\textwidth}{@{}
  >{\raggedright\arraybackslash}p{3.65cm}
  >{\raggedright\arraybackslash}X
  >{\raggedright\arraybackslash}p{2.65cm}@{}}
\toprule
\textbf{Type and definition} & \textbf{Example dialog} & \textbf{Annotation} \\
\midrule
\multicolumn{3}{@{}l}{\textit{Factuality hallucination}} \\
\midrule
\textbf{Factual Contradiction}\par States facts that can be grounded in real-world knowledge but contradict it, through a wrong entity or~relation.
  & \textit{Instruction}: How does the EU work?\par\smallskip \textit{Response}: The EU also has its own currency, the euro, which is used by \textbf{19} member states.
  & FAVABench, entity error; the correct number is 20. \\
\midrule
\textbf{Factual Fabrication}\par States facts that cannot be verified against established knowledge, or overclaims without broad support.
  & \textit{Instruction}: Name three important inventions of the 20th century.\par\smallskip \textit{Response}: The three important inventions of the 20th century are the airplane, the automobile, and \textbf{the chromoflare device, which enabled seamless digital color transitions in electronic displays}.
  & HADTest, fabricated entity; no such device exists. \\
\midrule
\multicolumn{3}{@{}l}{\textit{Faithfulness hallucination}} \\
\midrule
\textbf{Context Inconsistency}\par Is unfaithful to the context supplied by the user, such as a retrieved document.
  & \textit{Retrieved context}: Three of the alleged attackers were killed, Israeli media reported, citing IDF sources. It was not immediately known to what militant group the men belonged.\par\smallskip \textit{Instruction}: Summarize the following news within 38 words.\par\smallskip \textit{Response}: Israeli jets struck targets in the occupied Golan Heights after militants attempted to plant a bomb near the Israeli border. \textbf{Four alleged attackers were killed}, but their affiliation remains unclear.
  & RAGTruth, evident conflict; the source reports three deaths. \\
\midrule
\textbf{Logical Inconsistency}\par Contains internal contradictions, between statements or reasoning steps, or between the steps and the conclusion.
  & \textit{Document}: The text of this novel of ideas presents itself as a book that has been written as the result of a promise to a dying man. William Porphyry Benham is a man who has lived a life devoted to a complicated, protean idea: ``that he had to live life nobly and thoroughly.'' The novel has been written as \textbf{the result of a bet between two friends}.
  & ContraDoc; the last sentence contradicts the first. \\
\midrule
\textbf{Instruction Inconsistency}\par Deviates from an explicit, non-malicious \mbox{user instruction.}
  & \textit{Instruction}: what concentration of particles is bad for biosensors. The output should not contain any whitespace.\par\smallskip \textit{Response}: \textbf{Particle concentrations exceeding 1mg/mL or 10\textsuperscript{7}particles/mL often cause biosensor interference depending on sensor type particle size and matrix.}
  & IF-RewardBench; whitespace violates the instruction. \\
\bottomrule
\end{tabularx}\par
\captionof{table}{Hallucination taxonomy with examples from our benchmark. Definitions follow \citet{huang-etal-2025-survey}. Each example is an $H$ item; hallucinated content is in bold.}
\label{tab:taxonomy_examples}
\endgroup
\end{minipage}
\vspace{8pt}
}]

Table~\ref{tab:taxonomy_examples} lists the five hallucination types, their definitions, and benchmark examples.

\appsubsection{Dataset Versions and Construction}
\label{app:benchmark-details}
\begin{table*}[t]
\begingroup\centering\footnotesize
\setlength{\tabcolsep}{4pt}\renewcommand{\arraystretch}{1.15}
\begin{tabularx}{\textwidth}{@{}>{\raggedright\arraybackslash}p{2.25cm}>{\raggedright\arraybackslash}p{3.65cm}>{\raggedright\arraybackslash}X>{\raggedleft\arraybackslash}p{1.4cm}@{}}
\toprule
Dataset & Version / source snapshot & Split and selection & Instances \\
\midrule
FAVABench & \texttt{annotations.json}; SHA \texttt{ef3b4b289762} & All 460 human-annotated passages; separate inclusive FC and FF views. & FC: 377\newline FF: 351 \\
HADTest & \texttt{hadtest.json}; SHA \texttt{fd13225bcc8d} & Human-validated HADTest split only; HAD train and dev are not used. & FC: 1,244\newline FF: 1,265 \\
RAGTruth & Revision \texttt{c103204b9ce2} & Official \texttt{test} split, with \texttt{quality="good"}; join responses to source context by \texttt{source\_id}. & 2,675 \\
ContraDoc & Revision \texttt{d1c42d1e4335} & Entire released \texttt{pos}/\texttt{neg} benchmark; these are label partitions, not train/test splits. & 891 \\
ROSCOE & Official \texttt{annotations.zip}; SHA \texttt{1fe5b4ee56a6} & All released CosmosQA, DROP, e-SNLI, and GSM8K annotations; exclude one empty chain. & 755 \\
IF-RewardBench & Revision \texttt{c192fe968c15} & English-filtered released benchmark, including all three instruction forms. & 2,294 \\
\bottomrule
\end{tabularx}\par
\captionof{table}{Dataset versions and evaluation subsets. Counts are type-specific evaluation instances. No class balancing or additional train/test split is applied to these judge evaluations.}
\label{tab:dataset-versions}\par\endgroup
\end{table*}

Table~\ref{tab:dataset-versions} identifies the source snapshots, evaluation subsets, and resulting type-specific views. Revisions refer to the local source snapshots used to build the benchmark. Where an upstream revision was not retained, a SHA-256 fingerprint identifies the archived source file. Hashes below use the first 12 hexadecimal characters; the accompanying provenance file records the complete hashes and revision identifiers for the archived source snapshots.

\paragraph{Factuality mapping and reuse.}

For FAVABench, entity, relation, and contradictory map to FC; invented, unverifiable, and subjective map to FF, with subjective treated as overclaim. We reconstruct the annotated error-bearing passage by removing proposed corrections inside mark spans and stripping annotation tags while retaining their content, including delete spans. A response carrying both types is positive in both views. Negatives have none of the six factual error tags; responses carrying only the other type are excluded. Thus, 115 jointly positive and 153 jointly negative records are reused across FC and FF.

For HADTest, Factual Recall Error and Factual Inference Error map to FC, while Fabricated Entity and Fictional Attribution map to FF. The 1,124 None-labeled responses are negatives in both views; other categories are excluded. The conversion removes empty responses and conflicting-label duplicate texts, retaining the first occurrence of a same-label duplicate within a view. Reuse across types is intentional because the two tasks ask different questions of the same response.

\paragraph{Faithfulness subsets.}

RAGTruth is positive when at least one hallucination span is annotated, after the test/\texttt{good} filtering. ContraDoc includes both inserted and replacement-based contradictions. ROSCOE uses the whole-chain \texttt{0\_\allowbreak{}full\_\allowbreak{}newClear\allowbreak{}Contradiction\_\allowbreak{}result} label (\texttt{yes}/\texttt{no}); this annotation also mentions incoherent reasoning, so the view is broader than strict contradiction. Its 756 released rows yield 755 evaluable chains after excluding one empty CosmosQA response. IF-RewardBench's English filter retains instructions and checklists containing no characters in \texttt{U+4E00--U+9FFF}; it is a preprocessing heuristic, not a separate official language split. The retained responses comprise 984 single-turn, 574 multi-turn, and 736 system-prompt items. Any violated constraint maps to $H$.

\paragraph{Counting convention.}

The total of 9,852 counts \emph{type-specific evaluation instances}, not deduplicated answers. The 268 shared FAVABench records and 1,124 shared HADTest records account for the cross-type reuse. There are 8,460 distinct (dataset, source-record ID) pairs; this is an ID count, not global text deduplication. Judges receive complete responses, documents, or reasoning chains, without gold labels, annotated spans, or corrections.

\appsubsection{Binary Label Mapping}
\label{app:binary-mapping}
\begin{figure*}[t]
\centering
  \includegraphics[width=0.9\textwidth]{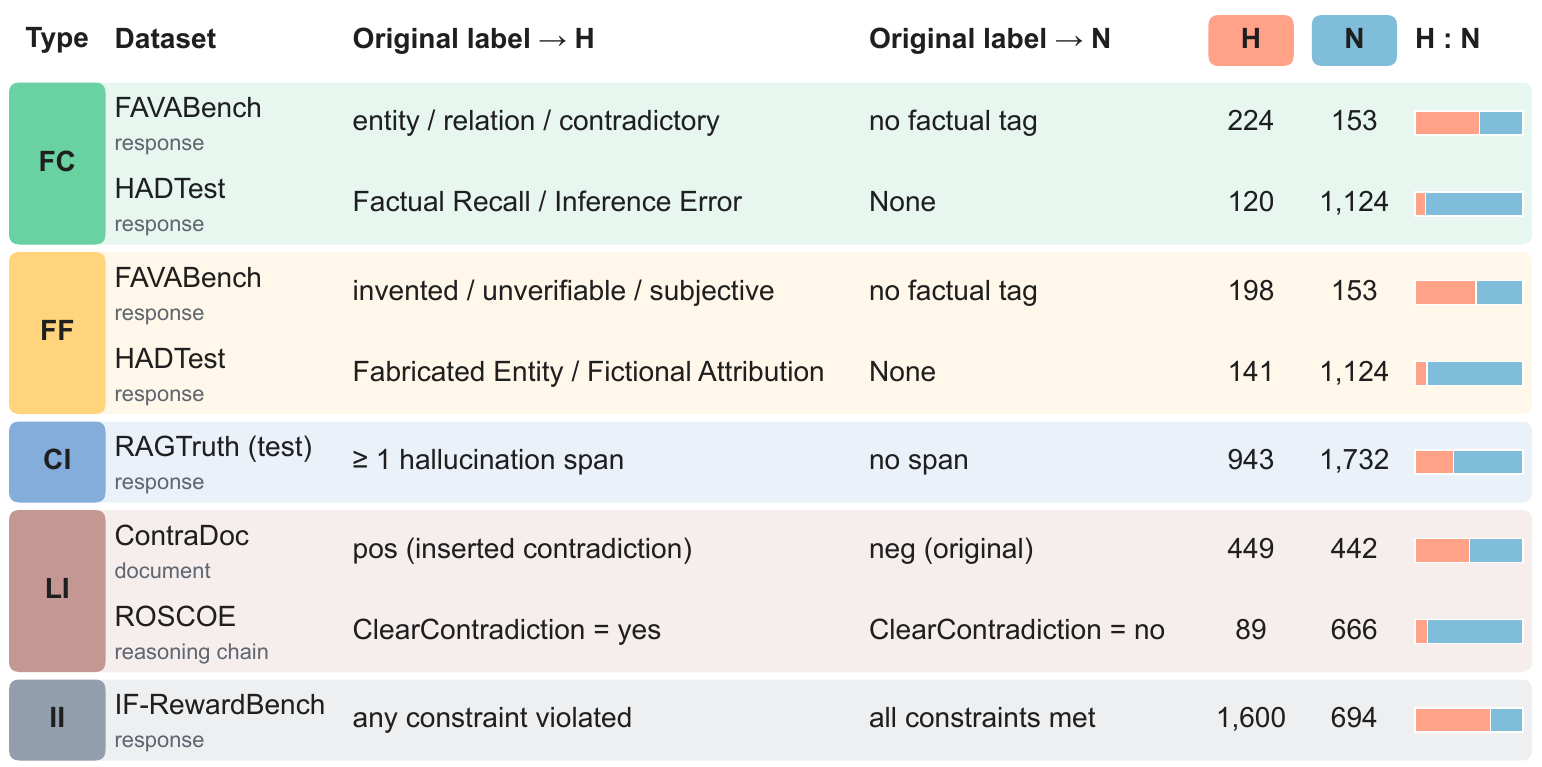}\captionof{figure}{Mapping of source-dataset annotations to binary labels. $H$: the target error is present; $N$: it is absent. The bar shows each dataset's $H{:}N$ ratio.}
  \label{fig:dataset_relabel}
  \vspace{-2mm}
\end{figure*}

Figure~\ref{fig:dataset_relabel} shows how we map each dataset's original annotations to binary labels. For each hallucination type, an item is labeled $H$ if its annotations indicate the target error and $N$ if they indicate its absence. FAVABench and HADTest provide annotations for both factual types, so we construct separate FC and FF views from each. In FAVABench, a response is negative only if it carries none of the six factual error tags. Responses tagged only with the other factual type are excluded. In HADTest, \texttt{None}-labeled responses serve as negatives for both FC and FF, and we exclude its other error categories. RAGTruth labels a response as $H$ if it contains at least one annotated hallucination span. ContraDoc uses its official split, and ROSCOE uses its whole-chain \texttt{ClearContradiction} label. In IF-RewardBench, a response is labeled $H$ if it violates any annotated constraint. We keep each dataset's original evaluation unit. Because some annotations target a single error type, $N$ means only that the target error is absent, not that the output \mbox{is error-free.}

\appsubsection{Decision-Model Prompts}
\label{app:dm-prompts}

Decision models have no separate system and user prompts. Each request carries a choice question, which is fixed for a hallucination type, and an item \texttt{state}, which is filled for each record. The decision models (jev, d1, clef, and pplx) receive both parts; gpt-6-luna receives the same text through the OpenAI Decisions API, with the \texttt{state} serialized as a JSON string and the $H$/$N$ criteria given as choice descriptions. Bracketed values are placeholders for the item text. A \texttt{state} field is sent only when it is non-empty. The shared request structure is shown below. The type-specific instructions and criteria are reproduced verbatim below; field headings are added for display, and the state fields retain the original placeholders.

\nolinenumbers
\begin{promptbox}[unbreakable]{Request structure shared by all types}
\begingroup\ttfamily\footnotesize\raggedright\setlength{\parindent}{0pt}
\noindent\{\newline
\hspace*{0.5em}"questions": \{\newline
\hspace*{1.0em}"target\_hallucination": \{\newline
\hspace*{1.5em}"type": "choice",\newline
\hspace*{1.5em}"instructions": "[Instructions]",\newline
\hspace*{1.5em}"criteria": \{\newline
\hspace*{2.0em}"H": "[Criterion H]",\newline
\hspace*{2.0em}"N": "[Criterion N]"\}\}\},\newline
\hspace*{0.5em}"state": \{[State fields]\}\newline
\}\par
\endgroup
\end{promptbox}

\pdfbookmark[3]{Factual Contradiction (FC)}{dm-0}
\begin{promptbox}{Prompt for factual contradiction (FC)}
\dmfield{Instructions}{Review the entire candidate response. Determine whether it contains factual contradiction. Choose H if at least one factual claim satisfies the H criterion. Otherwise choose~N. Evaluate only factual~contradiction.}
\dmfield{Criterion H}{At least one factual claim conflicts with reliable, verifiable real-world information. This includes an incorrect entity, relation, date, number, location, attribute, event, or other factual statement. Reliable evidence must establish that the claim is false.}
\dmfield{Criterion N}{No factual claim conflicts with reliable, verifiable real-world information. A claim that is merely unsupported, uncertain, subjective, or difficult to verify does not count as factual contradiction.}
\dmfield{State}{\texttt{candidate\_response}: "[the model-generated answer being judged]";\allowbreak \texttt{instruction}: "[the original user question or~prompt]"}
\end{promptbox}
\label{app:dm-factuality-templates}

\pdfbookmark[3]{Factual Fabrication (FF)}{dm-1}
\begin{promptbox}{Prompt for factual fabrication (FF)}
\dmfield{Instructions}{Review the entire candidate response. Determine whether it contains factual fabrication. Choose H if at least one factual claim satisfies the H criterion. Otherwise choose~N. Evaluate only factual~fabrication.}
\dmfield{Criterion H}{At least one statement is presented as established fact and satisfies at least one of the following conditions: (1) Unverifiability: it presents an invented or nonexistent entity, event, source, study, quotation, or other concrete factual detail as established fact without a reliable basis in real-world knowledge; or makes a concrete factual claim that cannot be substantiated by available reliable sources; (2) Overclaim: it presents a subjective, disputed, speculative, or exaggerated assertion as a broadly established fact despite lacking substantial supporting evidence or broad consensus. An incorrect entity, identity, attribute, relation, date, or number is not factual fabrication if the error solely contradicts an established fact.}
\dmfield{Criterion N}{The response contains no claim that is entirely nonexistent, unverifiable using available reliable sources, or an overclaim lacking broad consensus or substantial evidence. Choose N when all factual claims are verifiable, or when an error consists only of an incorrect entity substitution or relation that can be directly checked against established real-world facts.}
\dmfield{State}{\texttt{candidate\_response}: "[the model-generated answer being judged]";\allowbreak \texttt{instruction}: "[the original user question or~prompt]"}
\end{promptbox}

\pdfbookmark[3]{Context Inconsistency (CI)}{dm-2}
\begin{promptbox}{Prompt for context inconsistency (CI)}
\dmfield{Instructions}{Review the entire candidate response against the supplied context. Determine whether the response asserts at least one substantive factual claim that is inconsistent with or unsupported by that context. Choose H only if such a claim satisfies the H criterion. Otherwise choose~N. Use the supplied context as the evidence source, and evaluate only contextual faithfulness rather than formatting or general instruction compliance.}
\dmfield{Criterion H}{At least one substantive factual claim in the candidate response contradicts the supplied context or introduces information that cannot reasonably be supported or inferred from the supplied context. Judge faithfulness to the context even if the introduced information might be true in the real world.}
\dmfield{Criterion N}{Every substantive factual claim in the candidate response is supported by, entailed by, or reasonably inferable from the supplied context. Also choose N when the response merely omits contextual details, declines or is unable to answer, uses harmless introductory or meta wording, or expresses a reasonable paraphrase, summary, abstraction, or inference. Formatting, length, style, and other instruction-following violations do not by themselves count as context inconsistency.}
\dmfield{State}{\texttt{candidate\_response}: "[the model output being~judged]";\allowbreak \texttt{context}: "[the RAGTruth source material]";\allowbreak \texttt{instruction}: "[the RAGTruth task prompt]"}
\end{promptbox}
\label{app:dm-faithfulness-templates}

\pdfbookmark[3]{Logical Inconsistency (LI)}{dm-3}
\begin{promptbox}{Prompt for logical inconsistency (LI)}
\dmfield{Instructions}{Review the entire candidate response, including the relationships among its statements, reasoning steps, calculations, and conclusion. Determine whether it is internally logically inconsistent. Choose H if it contains any internal contradiction defined by the H criterion. Otherwise choose~N.}
\dmfield{Criterion H}{Two or more parts of the response are mutually incompatible; the response contradicts one of its earlier statements; or its conclusion conflicts with its own premises, assumptions, calculations, or stated reasoning steps.}
\dmfield{Criterion N}{The statements, reasoning steps, calculations, and conclusion are internally consistent with one another. A factual error, unverifiable claim, lack of contextual support, or failure to follow an instruction does not by itself count as internal logical~inconsistency.}
\dmfield{State}{\texttt{candidate\_response}: "[ContraDoc document | ROSCOE reasoning chain]";\allowbreak \texttt{instruction}: "Determine whether the document is self-contradictory." | "Determine whether the response is internally logically~inconsistent."}
\end{promptbox}

\pdfbookmark[3]{Instruction Inconsistency (II)}{dm-4}
\begin{promptbox}{Prompt for instruction inconsistency (II)}
\dmfield{Instructions}{Review the entire user instruction and candidate response. Determine whether the response fails to follow the instruction. Choose H if the response violates at least one applicable explicit requirement defined by the H criterion. Otherwise choose~N. Evaluate only instruction~compliance.}
\dmfield{Criterion H}{The response performs the wrong task, violates an explicit requirement or prohibition, omits required content or a required deliverable, or fails to satisfy an explicitly required format, language, length, style, role, structure, or other constraint.}
\dmfield{Criterion N}{The response performs the requested task and satisfies every applicable explicit requirement and constraint. Optional preferences do not have to be satisfied. A factual error, unverifiable statement, contextual error, or internal logical problem does not by itself count unless it directly violates an explicit instruction.}
\dmfield{State}{\texttt{candidate\_response}: "[the model response being judged]";\allowbreak \texttt{instruction}: "[the full IF-RewardBench prompt with its explicit constraints]"}
\end{promptbox}

\appsection{Model Configurations, Cost, and~Latency}
\label{app:model-config}
\noindent\begin{minipage}{\columnwidth}
\begingroup\centering\footnotesize
\setlength{\tabcolsep}{3pt}\renewcommand{\arraystretch}{1.2}
\begin{tabularx}{\columnwidth}{@{}>{\raggedright\arraybackslash}p{2.6cm}>{\raggedright\arraybackslash}p{1.7cm}>{\raggedright\arraybackslash}p{0.6cm}>{\raggedright\arraybackslash}X@{}}
\toprule
Model & Provider & Size & Version \\
\midrule
Jev & \mbox{TypeSafe AI} & NR & 1.13.0 \\
d1 & \mbox{Liquid AI} & NR & (not reported) \\
clef & \mbox{Cloudflare} & 27B & (not reported) \\
pplx-decider-\allowbreak{}v1.1-27b & \mbox{Perplexity} & 27B & v1.1 \\
GPT-6 Luna & \mbox{OpenAI} & NR & (not reported) \\
\midrule
GPT-6 Sol & \mbox{OpenAI} & NR & (not reported) \\
\bottomrule
\end{tabularx}\par
\captionof{table}{Model names, providers, sizes, and versions. The abbreviation pplx denotes pplx-decider-v1.1-27b. The first five rows are decision models; the final row is the generative reference. B denotes billion parameters; NR means the parameter count is not reported in the consulted official documentation for the evaluated API model. Published sizes follow the official \href{https://huggingface.co/Cloudflare/clef}{clef} and \href{https://huggingface.co/perplexity-ai/pplx-decider-v1.1-27b}{pplx} model cards.}
\label{tab:model-config}\par\endgroup
\end{minipage}
\par\vspace{8pt}
Table~\ref{tab:model-config} summarizes the providers, parameter counts, and recorded versions of the five decision models and the generative reference. The five decision models come from five model providers; OpenRouter is the hosting interface used for pplx, not its model provider.

Versions are extracted from the recorded model IDs; the date in the pplx row is the recorded snapshot suffix. For the other entries marked (not reported), the logs contain only an unversioned model alias and do not identify a specific checkpoint. The evaluated Liquid AI API model is d1; we do not assign it the parameter count of a separately released open-weight d1 variant.

\paragraph{Cost and latency.}
Table~\ref{tab:model-cost-latency} summarizes total estimated cost and mean API latency over all 9,852 type-specific evaluation instances per model. Cost is aggregated across the five types; latency is the mean of the recorded per-call times.

\noindent\begin{minipage}{\columnwidth}
\begingroup\centering\small
\setlength{\tabcolsep}{4pt}\renewcommand{\arraystretch}{1.2}
\begin{tabularx}{\columnwidth}{@{}Xrr@{}}
\toprule
Model & \shortstack{Total cost\\(USD)} & \shortstack{Mean latency\\(ms)} \\
\midrule
jev & 0.5409 & 271.22 \\
d1 & 0.4084 & 328.90 \\
clef & 2.5894 & 508.72 \\
pplx & 0.2092 & 306.14 \\
gpt-6-luna & 1.0393 & 256.26 \\
\midrule
gpt-6-sol & 30.7962 & 2,096.90 \\
\bottomrule
\end{tabularx}\par
\captionof{table}{Cost and latency across all evaluated instances. Cost sums the per-type estimates in Table~\ref{tab:model-performance}; latency averages recorded per-call times, weighted by the number of instances in each type. The cost estimator counts input tokens for decision models and both input and output tokens for gpt-6-sol.}
\label{tab:model-cost-latency}\par\endgroup
\end{minipage}

\appsection{Per-Dataset Results}
\label{app:rq1-details}
\begin{table*}[t]
\begingroup
\centering
\apptablestyle
\begin{tabular*}{\textwidth}{@{\extracolsep{\fill}}lrrrrrrrr@{}}
\toprule
Model & FC/FAVA & FC/HAD & FF/FAVA & FF/HAD & CI/RAGT. & LI/CDoc & LI/ROSC. & II/IFRB \\
\midrule
jev & 66.6 & 75.0 & 70.6 & 87.3 & 82.5 & 68.5 & 57.7 & 61.7 \\
d1 & 58.4 & 72.8 & 54.9 & 87.6 & 75.8 & 58.9 & 59.7 & 62.5 \\
clef & 62.7 & 72.7 & 57.7 & 85.4 & 68.2 & 62.9 & 64.5 & 64.5 \\
pplx & 65.2 & 76.8 & 63.8 & 94.8 & 82.7 & 75.0 & 67.2 & 71.1 \\
gpt-6-luna & 65.6 & 73.0 & 68.5 & 86.3 & 64.0 & 68.2 & 61.1 & 54.2 \\
gpt-6-sol & 67.3 & 80.4 & 68.4 & 92.6 & 82.7 & 77.7 & 65.2 & 78.9 \\
\midrule
$n$ ($H$) & 377 (224) & 1244 (120) & 351 (198) & 1265 (141) & 2675 (943) & 891 (449) & 755 (89) & 2294 (1600) \\
\bottomrule
\end{tabular*}
\par
\captionof{table}{BAcc (\%) per source dataset. FAVA is FAVABench, HAD is HADTest, RAGT.\ is RAGTruth, CDoc is ContraDoc, ROSC.\ is ROSCOE, and IFRB is IF-RewardBench.}
\label{tab:rq1-per-dataset}\par
\endgroup
\par\vspace{14pt}
\begingroup
\centering
\apptablestyle
\begin{tabular*}{\textwidth}{@{\extracolsep{\fill}}lrrrrrrrr@{}}
\toprule
Model & FC/FAVA & FC/HAD & FF/FAVA & FF/HAD & CI/RAGT. & LI/CDoc & LI/ROSC. & II/IFRB \\
\midrule
jev & 38/29 & 42/8 & 25/34 & 5/20 & 21/14 & 49/14 & 81/4 & 20/56 \\
d1 & 59/24 & 47/8 & 80/10 & 23/1 & 41/7 & 81/2 & 76/4 & 26/49 \\
clef & 48/27 & 50/5 & 74/10 & 28/2 & 60/3 & 73/1 & 62/9 & 22/49 \\
pplx & 28/42 & 41/6 & 56/16 & 8/3 & 31/3 & 45/5 & 58/7 & 41/17 \\
gpt-6-luna & 5/63 & 27/27 & 15/48 & 9/19 & 2/70 & 56/8 & 70/8 & 3/89 \\
gpt-6-sol & 17/48 & 28/12 & 41/22 & 8/7 & 6/29 & 31/13 & 63/7 & 14/28 \\
\bottomrule
\end{tabular*}
\par
\captionof{table}{False-negative rate / false-positive rate (\%) per source dataset.}
\label{tab:rq1-fnr-fpr}\par
\endgroup
\end{table*}

Table~\ref{tab:rq1-per-dataset} reports BAcc for every judge on each source dataset, and Table~\ref{tab:rq1-fnr-fpr} the corresponding false-negative and false-positive rates. \par

\appsection{Error Analysis}
\label{app:error-analysis}
\appsubsection{Errors by Source Annotation}
\begin{table*}[t]
\begingroup\centering\apptablestyle
\begin{tabular*}{\textwidth}{@{\extracolsep{\fill}}llrrrrr@{}}
\toprule
Attribute (dataset, type) & Level & $n$ & pplx & jev & gpt-6-luna & gpt-6-sol \\
\midrule
\multicolumn{7}{@{}l}{\textit{Error density}} \\
Annotated spans (RAGTruth, CI) & 1 & 590 & 57 & 71 & 97 & 91 \\
 & 2 & 206 & 81 & 90 & 99 & 99 \\
 & $\geq$3 & 147 & 96 & 98 & 100 & 99 \\
Violated constraints (IF-RewardBench, II) & 1 & 764 & 46 & 73 & 95 & 76 \\
 & 2 & 465 & 68 & 83 & 98 & 91 \\
 & $\geq$3 & 371 & 74 & 89 & 100 & 98 \\
\midrule
\multicolumn{7}{@{}l}{\textit{Salience}} \\
Span type (RAGTruth, CI) & evident & 866 & 71 & 82 & 98 & 95 \\
 & subtle only & 77 & 42 & 55 & 97 & 78 \\
Contradiction plug (ContraDoc, LI) & insert & 178 & 91 & 84 & 73 & 93 \\
 & replace & 271 & 32 & 29 & 25 & 53 \\
Contradiction type (ContraDoc, LI) & negation / numeric & 63 & 86 & 86 & 83 & 90 \\
 & content / factual & 306 & 55 & 50 & 41 & 71 \\
 & perspective / emotion & 58 & 31 & 26 & 28 & 40 \\
Fabrication subtype (FAVABench, FF) & invented & 48 & 67 & 85 & 100 & 71 \\
 & subjective / unverifiable & 138 & 35 & 70 & 80 & 54 \\
\midrule
\multicolumn{7}{@{}l}{\textit{Task form (BAcc)}} \\
Task type (RAGTruth, CI) & QA & 875 & 81 & 82 & 65 & 81 \\
 & Summary & 900 & 72 & 74 & 66 & 82 \\
 & Data2txt & 900 & 86 & 82 & 51 & 79 \\
Instruction type (IF-RewardBench, II) & single turn & 984 & 77 & 66 & 58 & 79 \\
 & multi turn & 574 & 63 & 53 & 51 & 72 \\
 & system prompt & 736 & 68 & 59 & 52 & 85 \\
Reasoning source (ROSCOE, LI) & gsm8k & 200 & 81 & 67 & 76 & 84 \\
 & cosmos & 194 & 57 & 52 & 53 & 53 \\
 & drop & 210 & 63 & 52 & 51 & 61 \\
 & esnli & 151 & 68 & 49 & 49 & 48 \\
Document type (ContraDoc, LI) & wiki & 295 & 83 & 77 & 74 & 80 \\
 & news & 312 & 76 & 68 & 71 & 82 \\
 & story & 284 & 65 & 60 & 59 & 71 \\
\midrule
\multicolumn{7}{@{}l}{\textit{Generator (BAcc)}} \\
Response generator (RAGTruth, CI) & GPT-3.5 / GPT-4 & 891 & 54 & 66 & 66 & 80 \\
 & Llama-2 (7B/13B/70B) & 1348 & 85 & 82 & 62 & 78 \\
 & Mistral-7B & 436 & 85 & 79 & 65 & 81 \\
\bottomrule
\end{tabular*}\par
\captionof{table}{Where judges fail. Rows under \textit{Error density} and \textit{Salience} report recall of $H$ (\%) on the $H$ items with the given attribute. Rows under \textit{Task form} and \textit{Generator} report BAcc (\%) on all items with the attribute. Attributes come from the source annotations and were not shown to the judges.}
\label{tab:rq1-difficulty}\par
\endgroup
\end{table*}

Table~\ref{tab:rq1-difficulty} breaks recall of $H$ down by attributes available in the source annotations. The attributes are the number of annotated error spans or violated constraints, the salience of the error, the task form, and the model that generated the judged response. Where both classes are present, the table reports BAcc instead of recall.

\appsubsection{Error Patterns and Overlap}
\label{app:error-patterns}
\paragraph{Errors concentrate on sparse and subtle targets.}
To locate where decision models fail, we break recall of $H$ down by the source annotations (Table~\ref{tab:rq1-difficulty}). The amount of evidence matters most. On RAGTruth, \textit{pplx} recovers 57\% of responses with one annotated span but 96\% of those with three or more. On IF-RewardBench, it recovers 46\% of responses that violate one constraint but 74\% of those that violate three or more. \textit{gpt-6-sol} recovers 91\% and 76\% of the single-error cases. Salience matters next. The recall of \textit{pplx} drops from 71\% to 42\% for RAGTruth spans annotated only as subtle, and from 91\% to 32\% for ContraDoc contradictions created by replacing a sentence rather than inserting one. Task form matters as well. All decision models fall to 47 to 63\% BAcc on multi-turn instructions, and every judge, including \textit{gpt-6-sol}, is close to chance on the non-arithmetic ROSCOE chains. The generative judge is not uniformly better either. On RAGTruth data-to-text items, \textit{pplx} reaches 86\% BAcc against 79\% for \textit{gpt-6-sol}.

\paragraph{Decision-model errors overlap only partly with the generative judge.}
\textit{pplx} and \textit{gpt-6-sol} are both wrong on 4\% of CI items, 6 to 8\% of FC and FF items, and 11 to 13\% of LI and II items. The Jaccard overlap of their error sets rises from 0.13 on CI to 0.50 on LI. Overlap among decision models is higher and is again largest on LI, between 0.51 and 0.57 for \textit{pplx} against each other decision model. Errors are thus least shared on the types where decision models are already strong and most shared where they are weak. This pattern anticipates RQ3.

\appsection{Confidence Analysis Details}
\label{app:conf-details}
\begin{figure*}[t]
\centering
  \includegraphics[width=0.83\textwidth]{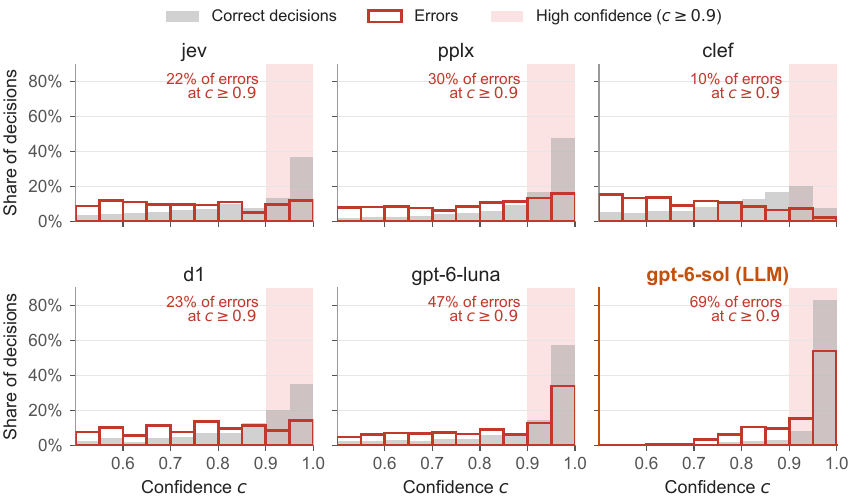}\captionof{figure}{\textbf{Confidence distributions of correct decisions and errors.} For each model, the share of correct decisions (gray bars) and of errors (red outline) in each confidence bin. The results are averaged over the five hallucination types, with each dataset weighted equally within its type.}
  \label{fig:conf-dist}
\end{figure*}
\paragraph{Confidence and hallucination type.}
To test whether the accuracy implied by a confidence level depends on the hallucination type $t$, we fit, for each decision model, $\operatorname{logit}\Pr(\hat{y}_i = y_i) = \beta_0 + \beta_1 c_i + \gamma_t + \delta_t c_i$, with standard errors clustered by response, and test $\delta_t = 0$ for all $t$. The interaction is significant for every model (Wald $\chi^2 = 57$--$142$, all $p < 10^{-10}$).

\paragraph{Confidence distributions.}
Figure~\ref{fig:conf-dist} shows, for each model, how its correct decisions and its errors are distributed over confidence. For a useful confidence score, errors should concentrate at low confidence and correct decisions at high confidence. This holds most clearly for \textit{clef}, whose errors are spread over the lower confidence range and only 10\% of which fall at $c \geq 0.9$. For \textit{gpt-6-luna} and especially \textit{gpt-6-sol}, the error distribution peaks in the highest confidence bin, much like that of correct decisions, so high confidence alone does not reliably distinguish between the two.

\paragraph{Robustness check of the calibration results.}
The ECE ranking of the models is unchanged under 15 equal-width bins, under 10 or 15 equal-mass bins, and under a debiased ECE. Clipping probabilities at $10^{-2}$ instead of $10^{-4}$ changes each model's median calibration slope by at most $0.09$.

\appsection{Cascade Deferral-Rate Sensitivity}
\label{app:cascade-sensitivity}
\begin{figure*}[tp]
\centering
  \includegraphics[width=\textwidth]{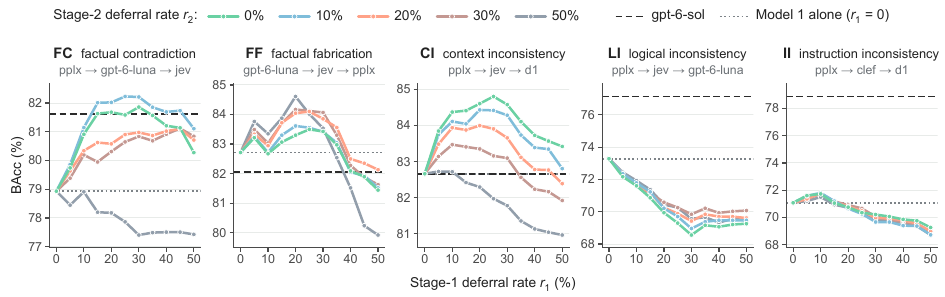}\captionof{figure}{\textbf{Cascade performance across deferral rates.} The x-axis varies $r_1$; each colored line fixes $r_2$. Both rates range from 0\% to 50\% in 10-percentage-point steps (36 configurations per type). Stars mark the highest BAcc, dashed lines mark gpt-6-sol, and dotted lines mark the strongest member alone. Members appear in stage order below each panel. Peaks are selected and evaluated on the same items; they are post-hoc maxima, not held-out~results.}
  \label{fig:cascade}
\end{figure*}

Figure~\ref{fig:cascade} shows sensitivity to two target deferral rates: $r_1$ is the fraction of all items passed from the first member to the second, and $r_2$ is the fraction of second-stage items passed to the third. Each stage defers its least-confident items. At $r_1=0$, all curves coincide with the strongest member alone.

\paragraph{Moderate deferral helps selected types.}

FC, FF, and CI peak at intermediate $r_1$ values (20--30\%), reaching 82.2\%, 84.6\%, and 84.6\% BAcc, respectively; larger rates eventually reduce performance. The separation between curves also shows sensitivity to $r_2$, particularly for FC and CI. LI performs best without deferral: every tested nonzero $r_1$ lowers BAcc. II gains at most 0.7 percentage points over the strongest member and remains 7.1 points below gpt-6-sol. Thus, deferring more low-confidence items does not consistently improve accuracy; the benefit depends on the hallucination type and both operating rates.

\appsubsection{Peak and Cross-Fitted Results}
\label{app:cascade-results}
\begin{table*}[p]
\begingroup\centering\footnotesize
\setlength{\tabcolsep}{4pt}\renewcommand{\arraystretch}{1.0}
\begin{tabular*}{\textwidth}{@{\extracolsep{\fill}}llrrrr@{}}
\toprule
Type & Evaluation & BAcc & $\Delta$ single [95\% CI] & $\Delta$ sol [95\% CI] & $\Delta$ peak \\
\midrule
FC & Peak & 82.21 & $+3.28$ [0.98, 5.57] & $+0.60$ [-1.88, 2.96] & $+0.00$ \\
FC & Cross-fitted & 81.69 & $+2.76$ [0.63, 4.85] & $+0.08$ [-2.33, 2.51] & $-0.52$ \\
\addlinespace[1pt]
FF & Peak & 84.61 & $+1.89$ [0.38, 3.45] & $+2.57$ [0.02, 5.14] & $+0.00$ \\
FF & Cross-fitted & 84.39 & $+1.67$ [0.19, 3.18] & $+2.34$ [-0.22, 4.76] & $-0.22$ \\
\addlinespace[1pt]
CI & Peak & 84.61 & $+1.94$ [0.65, 3.24] & $+1.95$ [0.38, 3.57] & $+0.00$ \\
CI & Cross-fitted & 84.24 & $+1.58$ [0.36, 2.84] & $+1.59$ [-0.07, 3.19] & $-0.37$ \\
\addlinespace[1pt]
LI & Peak & 73.28 & $+0.00$ [0.00, 0.00] & $-3.85$ [-5.89, -1.84] & $+0.00$ \\
LI & Cross-fitted & 73.28 & $+0.00$ [0.00, 0.00] & $-3.85$ [-5.82, -1.77] & $+0.00$ \\
\addlinespace[1pt]
II & Peak & 71.80 & $+0.72$ [-0.12, 1.60] & $-7.07$ [-9.38, -4.85] & $+0.00$ \\
II & Cross-fitted & 71.44 & $+0.36$ [-0.51, 1.20] & $-7.43$ [-9.71, -5.02] & $-0.36$ \\
\addlinespace[1pt]
\bottomrule
\end{tabular*}\par
\captionof{table}{BAcc and paired differences in percentage points; single is the first member alone. Cross-fitted BAcc averages five seeds. Intervals use 2,000 paired item-bootstrap resamples, sharing indices across seeds before averaging. They condition on selected members and recorded predictions; peak intervals are not selection-adjusted.}
\label{tab:cascade-comparison}\par\endgroup
\par\vspace{14pt}
\begingroup\centering\footnotesize

\setlength{\tabcolsep}{4pt}\renewcommand{\arraystretch}{1.0}
\begin{tabular*}{\textwidth}{@{\extracolsep{\fill}}llrrrrrr@{}}
\toprule
Type & Evaluation & Realized $r_1$ & Realized $r_2$ & USD/1k & Cost (\% sol) & Calls & Lat. (ms) \\
\midrule
FC & Peak & 30.0 & 9.3 & 0.0275 & 1.34 & 1.33 & 362.8 \\
FC & Cross-fitted & 27.0 & 10.0 & 0.0257 & 1.25 & 1.30 & 355.5 \\
FF & Peak & 19.7 & 49.2 & 0.0717 & 3.31 & 1.29 & 313.0 \\
FF & Cross-fitted & 20.8 & 48.0 & 0.0722 & 3.33 & 1.31 & 316.2 \\
CI & Peak & 20.0 & 0.0 & 0.0511 & 1.25 & 1.20 & 398.2 \\
CI & Cross-fitted & 22.6 & 0.0 & 0.0535 & 1.30 & 1.23 & 405.7 \\
LI & Peak & 0.0 & -- & 0.0164 & 0.60 & 1.00 & 284.8 \\
LI & Cross-fitted & 0.0 & -- & 0.0164 & 0.60 & 1.00 & 284.8 \\
II & Peak & 10.0 & 10.0 & 0.0576 & 1.55 & 1.11 & 371.2 \\
II & Cross-fitted & 10.1 & 12.9 & 0.0580 & 1.56 & 1.11 & 372.9 \\
\bottomrule
\end{tabular*}\par
\captionof{table}{Realized routing percentages and replay efficiency. Realized $r_2$ is conditional on reaching stage 2 (-- if none). Cross-fitted rates pool counts across seeds; other quantities average over items and seeds. Confidence ties can make realized rates differ from their targets.}
\label{tab:cascade-efficiency}\par\endgroup
\par\vspace{14pt}
\begingroup\centering\footnotesize
\setlength{\tabcolsep}{4pt}\renewcommand{\arraystretch}{1.0}
\begin{tabularx}{\textwidth}{@{}lp{4.35cm}lX@{}}
\toprule
Type & Members in stage order & Peak $(r_1,r_2)$ & Cross-fitted targets: setting $\times$ fold count \\
\midrule
FC & pplx $\rightarrow$ gpt-6-luna $\rightarrow$ jev & $(0.3,0.1)$ & $(0.2,0.1)\times 7$; $(0.3,0.1)\times 17$; $(0.4,0.1)\times 1$ \\
FF & gpt-6-luna $\rightarrow$ jev $\rightarrow$ pplx & $(0.2,0.5)$ & $(0.2,0.5)\times 23$; $(0.3,0.3)\times 2$ \\
CI & pplx $\rightarrow$ jev $\rightarrow$ d1 & $(0.2,0.0)$ & $(0.1,0.0)\times 2$; $(0.2,0.0)\times 14$; $(0.3,0.0)\times 9$ \\
LI & pplx $\rightarrow$ jev $\rightarrow$ gpt-6-luna & $(0.0,0.0)$ & $(0.0,0.0)\times 25$ \\
II & pplx $\rightarrow$ clef $\rightarrow$ d1 & $(0.1,0.1)$ & $(0.1,0.0)\times 5$; $(0.1,0.1)\times 10$; $(0.1,0.2)\times 6$; $(0.1,0.3)\times 4$ \\
\bottomrule
\end{tabularx}\par
\captionof{table}{Fixed member order and selected target rates. Cross-fitted counts cover 25 fits (five folds for each of five seeds). The target $r_2$ applies only to items reaching the second stage.}
\label{tab:cascade-settings}\par\endgroup
\end{table*}

\looseness=-1
Peak settings maximize pooled BAcc on the 36-point grid. Cross-fitting uses five label-stratified folds and seeds 0--4: fit rates and confidence thresholds on four folds, then apply them to the held-out fold. Members stay ordered by full-sample BAcc, so this checks rate/threshold selection, not the full model-selection pipeline. Ties use ascending $(r_1,r_2)$ order.

\looseness=-1
Cross-fitting lowers BAcc by at most 0.52 points. FC, FF, and CI retain gains over the strongest member, with intervals excluding zero. Their cross-fitted differences from gpt-6-sol have intervals containing zero. LI always selects no deferral; II has no clear gain over its strongest~member.






\end{document}